%% file: paper.tex
\pdfoutput=1

\documentclass[10pt,twocolumn,letterpaper]{article}

\usepackage{cvpr}
\usepackage{times}
\usepackage{epsfig}
\usepackage{graphicx}
\usepackage{amsmath}
\usepackage{amssymb}
\usepackage{multirow}
\usepackage{makecell}
\usepackage{amsmath, optidef}
\usepackage[caption=false]{subfig}

\usepackage{amsmath}

\usepackage[pagebackref=true,breaklinks=true,colorlinks,bookmarks=false]{hyperref}

\usepackage{times}
\usepackage{epsfig}
\usepackage{graphicx}
\usepackage{amsmath}
\usepackage{amssymb}
\usepackage{algorithm}
\usepackage{algorithmic}
\usepackage{adjustbox}
\usepackage{multirow}
\usepackage{makecell}
\usepackage{xcolor}

\newcommand{\deemph}[1]{{\color{black!40}#1}}
\newcommand{\featext}{\textsc{ExtractFeature}}
\newcommand{\roipool}{\textsc{RoIPool}}
\newcommand{\rpn}{\textsc{RPN}}
\cvprfinalcopy %
\def\cvprPaperID{3400} 

\begin{document}

\title{Tracking Pedestrian Heads in Dense Crowd}


\author{
  \hspace{-1.3cm}
  \begin{tabular}[t]{c}
    Ramana Sundararaman \quad C\'edric De Almeida Braga \quad Eric Marchand \quad Julien Pettr\'e\\
    Univ Rennes, Inria, CNRS, Irisa, Rennes, France \\
     {\tt\small ramanasubramanyam.sundararaman@polytechnique.edu, } \\
     {\tt\small \{cedric.de-almeida-braga,julien.pettre\}@inria.fr, }\\
     {\tt\small eric.marchand@irisa.fr}

\end{tabular}
}

\maketitle

\begin{abstract}
    
Tracking humans in crowded video sequences is an important constituent of visual scene understanding. Increasing crowd density challenges visibility of humans, limiting the scalability of existing pedestrian trackers to higher crowd densities. For that reason, we propose to revitalize head tracking with Crowd of Heads Dataset (CroHD), consisting of 9 sequences of 11,463 frames with over 2,276,838 heads and 5,230 tracks annotated in diverse scenes. For evaluation, we proposed a new metric, IDEucl, to measure an algorithm's efficacy in preserving a unique identity for the longest stretch in image coordinate space, thus building a correspondence between pedestrian crowd motion and the performance of a tracking algorithm. Moreover, we also propose a new head detector, HeadHunter, which is designed for small head detection in crowded scenes. We extend HeadHunter with a Particle Filter and a color histogram based re-identification module for head tracking. To establish this as a strong baseline, we compare our tracker with existing state-of-the-art pedestrian trackers on CroHD and demonstrate superiority, especially in identity preserving tracking metrics. With a light-weight head detector and a tracker which is efficient at identity preservation, we believe our contributions will serve useful in advancement of pedestrian tracking in dense crowds.

\end{abstract}

\section{Introduction}

Tracking multiple objects, especially humans, is a central problem in visual scene understanding. The intricacy of this task grows with increasing targets to be tracked and remains an open area of research. Alike other subfields in Computer Vision, with the advent of Deep Learning, the task of Multiple Object Tracking (MOT) has remarkably advanced its benchmarks~\cite{dave_tao,Geiger2013VisionMR,KITTI,MOTChallenge2015,MOT16,MOTS} since its inception~\cite{pets_dataset}. In the recent past, the focus of MOTChallenge benchmark~\cite{MOT19_CVPR} has shifted towards tracking pedestrians in crowds of higher density. This has several applications in fields such as activity recognition, anomaly detection, robot navigation, visual surveillance, safety planning etc.
\begin{figure}[t]
\begin{center}
\includegraphics[width=0.8\linewidth]{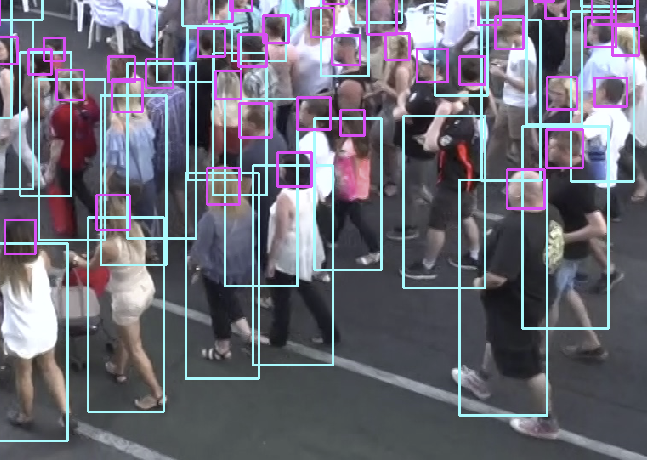}
\end{center}
\caption{Comparison between head detection and full body detection in a crowded scene from CroHD. HeadHunter detects 36 heads whereas Faster-RCNN~\cite{FRCNN} can detect only 23 pedestrians out of 37 present in this scene.}
\label{fig:bodyvshead}
\end{figure}
Yet, the performances of trackers on these benchmark suggests a trend of saturation\footnote{https://motchallenge.net/results/MOT20/}. Majority of online tracking algorithms today follow the tracking-by-detection paradigm and several research works have well-established object detector's performance to be crucial in tracker's performance~\cite{tracktor, SORT, tracking_survey}. As the pedestrian density in a scene increases, pedestrian visibility reduces with increasing mutual occlusions, leading to reduced pedestrian detection as visualized in Figure~\ref{fig:bodyvshead}. To tackle these challenges yet track humans efficiently in densely crowded environments, we rekindle the task of MOT with tracking humans by their distinctly visible part - heads. To that end, we propose a new dataset, \emph{CroHD, Crowd of Heads Dataset}, comprising 9 sequences of 11,463 frames with head bounding boxes annotated for tracking. We hope that this new dataset opens up opportunities for promising future research to better understand global pedestrian motion in dense crowds.

Supplementing this, we develop two new baseline methods on CroHD, a head detector, \emph{HeadHunter} and a head tracker, \emph{HeadHunter-T}. We design HeadHunter peculiar for head detection in crowded environments, distinct from standard pedestrian detectors and demonstrate state-of-the-art performance on an existing head detection dataset. HeadHunter-T extends HeadHunter with a Particle Filter framework and a light-weight re-identification module for head-tracking. To validate HeadHunter-T to be a strong baseline tracker, we compare it with three published top performing pedestrian trackers on the crowded MOTChallenge benchmark, evaluated on CroHD. We further perform comparisons between tracking by head detection and tracking by body detection to illustrate the usefulness of our contribution.

To establish correspondence between a tracking algorithm and pedestrian motion, it is necessary to understand the adequacy of various trackers in successfully representing ground truth pedestrian trajectories. We thus propose a new metric, \emph{IDEucl} to evaluate tracking algorithms based on their consistency in maintaining the same identity for the longest length of a ground truth trajectory in the image coordinate space. \emph{IDEucl} is compatible with our dataset and can be extended to any tracking benchmark, recorded with a static camera. \\
In summary, this paper makes the following contributions \textbf{(i)} We present a new dataset, CroHD, with annotated pedestrian heads for tracking in dense crowd, \textbf{(ii)} We propose a baseline head detector for CroHD, HeadHunter, \textbf{(iii)} We develop HeadHunter-T, by extending HeadHunter as the baseline head tracker for CroHD, \textbf{(iv)} We propose a new metric, IDEucl, to evaluate the efficiency of trackers in representing a ground truth trajectory and finally, \textbf{(v)} We demonstrate HeadHunter-T to be a strong baseline by comparing with three existing state-of-the-art trackers on CroHD.

\section{Related Work}

\textbf{Head Detection Benchmarks:} The earliest benchmarks in head detection are~\cite{TVHI,zisserman_head,HollywoodHeads,CasaBlanca}, which provide ground truth head annotations of subjects in Hollywood movies. In the recent past, SCUT-Head~\cite{scut_head} and CrowdHuman dataset~\cite{shao2018crowdhuman} provide head annotations of humans in crowded scenes. Head detection is also of significant interest in the crowd counting and analysis literature~\cite{idress_ccounting}. Rodriguez \etal~\cite{data_driven_crowd} introduced the idea of tracking by head detection with their dataset consisting of roughly 2200 head annotations. In the recent years, there has been a surge in research works attempting to narrow the gap between detection and crowd counting~\cite{decidenet,LSCCNN20,point_in_box,MCNN} which attempts to hallucinate pseudo head ground truth bounding boxes in crowded scenes.

\textbf{Head Detection Methods:} Fundamentally, the task of head detection is a combination of multi-scale and contextual object detection problem. Objects at multiple scales are detected based on image pyramids~\cite{tiny_face,scut_head,analysissnip2017,sniper2018,ext_tiny_face} or feature pyramids~\cite{maskrcnn,FPN,Small_Faces}. The former is computationally intensive task requiring multiple forward passes of images while the latter generates multiple pyramids in a single forward pass. Contextual object detection has been widely addressed in the literature of face detection, such as~\cite{retina_face,SSH,PyramidBox} who show improved detection accuracy by employing convolutional filters of larger receptive size to model context. Sun \etal~\cite{Sun_SMD} employ such a contextual and scale-invariant applied to head detection. 

\textbf{Tracking Benchmarks and Metrics:} The task of Multiple Object Tracking (MOT) is to track an initially unknown number of targets in a video sequence. The first MOT dataset for tracking humans were the PETS dataset~\cite{pets_dataset}, soon followed by \cite{TUD_seq,ETH,Geiger2013VisionMR,KITTI}. Standardization of MOT benchmarks were later proposed in \cite{MOTChallenge2015} and since then, it has been updated with yearly challenges involving more complex scenarios and increasingly crowded environments~\cite{MOT19_CVPR,MOT16}. Recently, the TAO dataset~\cite{dave_tao} was introduced for Multi-object tracking, which focuses on tracking 833 object categories across 2907 short sequences. Our dataset pushes the challenge of tracking in crowded environments with pedestrian density reaching 346 humans per frame. Other relevant pedestrian tracking dataset include \cite{USC,wildtrack_dataset,MOTS}. \\
To evaluate algorithms on MOTChallenge dataset, classical MOT metrics~\cite{classic_metric} and CLEAR MOT metrics~\cite{CLEARMetric} have been \textit{de facto} established as standardised way of quantifying performances. The CLEAR Metric proposes two important scores MOTA and MOTP which concisely summarise the classical metrics based on cumulative per frame accuracy and precision of bounding boxes respectively. Recently, Ristani \etal~\cite{DukeMTMC} propose the ID metric, which rewards a tracker based on its efficiency in preserving an identity for the longest duration of the Ground Truth trajectory. 

\begin{figure*}
\begin{center}
 \begin{minipage}{0.20 \textwidth}
\centering 
\textbf{Scene 1}
 \end{minipage}%
  \begin{minipage}{0.20 \textwidth}
\centering  \textbf{Scene 2}
 \end{minipage}%
  \begin{minipage}{0.20 \textwidth}
\centering \textbf{ Scene 3}
\end{minipage}%
\begin{minipage}{0.20 \textwidth}
\centering \textbf{ Scene 4}
 \end{minipage}%
 \begin{minipage}{0.20 \textwidth}
 \centering \textbf{ Scene 5}
 \end{minipage}
 \includegraphics[width=\linewidth]{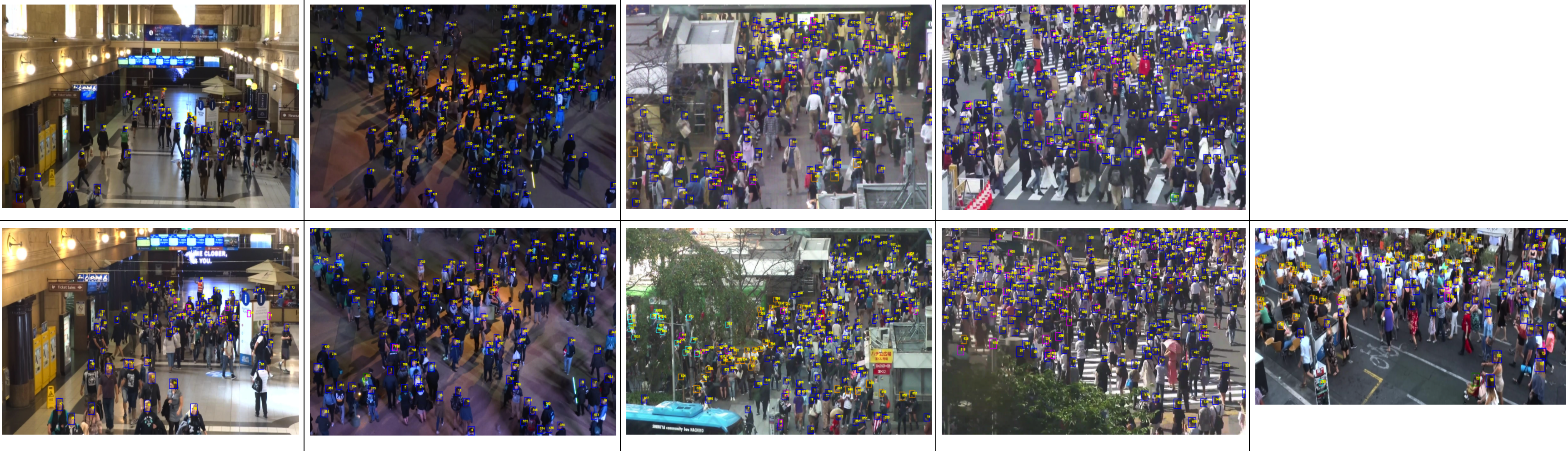}
 \end{center}
 \caption{Depiction of a frame per each scene from our Crowd of Heads Dataset, CroHD. The top row shows frames from the training set while the bottom row illustrates frames from the test set.}
 \label{fig:dataOverview}
\end{figure*}

\textbf{Tracking Algorithms:} Online Multi-object tracking algorithms can be summarized down into: (i) Detection, (ii) Motion Prediction, (iii) Affinity Computation and, (iv) Association steps. R-CNN based networks have been common choice for the detection stage due to the innate advantage of proposal based detectors over Single-Stage detection methods \cite{objdet_comparison}. Amongst online Multiple Object Tracking algorithm, Chen \etal~\cite{chen_pf} use Particle Filter framework and weigh the importance of each particle by their appearance classification score, computed by a separate network, trained independently. Earlier works such as \cite{breitenstein_iccv,xing_pf} use Sequential Importance Sampling (SIS) with Constant Velocity Assumption for assigning importance weight to particles. Henschel \etal~\cite{fusion_head} demonstrated the the limitation of single object detector for tracking and used a head detector~\cite{brainwash} in tandem with the pedestrian detector~\cite{FRCNN}. However, in the recent past, research works in MOT have attempted to bridge the gap between tracking and detection through a unified framework~\cite{tracktor, RANN,track_to_detect,Kieritz_jdomot,fast_slow,MOTS}. Most notable amongst them is Tracktor~\cite{tracktor}, who demonstrated that an object detector alone is sufficient to predict locations of targets in subsequent frames, benefiting from the high-frame rates in video. 


\section{CroHD Dataset}

\textbf{Description:}
The objective of CroHD is to provide tracking annotation of pedestrian heads in densely populated video sequences. To the best of our knowledge, no such benchmark exists in the community and hence we annotated 2,276,838 human heads in 11,463 frames across 9 sequences of Full-HD resolution. We built CroHD upon 5 sequences from the publicly available MOTChallenge CVPR19 benchmark~\cite{MOT19_CVPR} to enable performance comparison of trackers in the same scene between two paradigms - head tracking and pedestrian tracking. We maintain the training set and test set classification of the aforementioned sequences to be the same in CroHD as the MOTChallenge CVPR19 benchmark. We further annotated 4 new sequences of higher crowd densities in two new scenarios. The new scenario centers on the Shibuya Train station and Shibuya Crossing, one of the busiest pedestrian crossings in the world. All sequences in CroHD have a frame-rate of $25 fps$ and are captured from an elevated viewpoint. The sequences involve crowded indoor and outdoor scenes, recorded across different lighting and environmental conditions. This ensures sufficient diversity in the dataset in order to make it viable for training and evaluating the comprehensiveness of modern Deep Learning based techniques. The maximum pedestrian density reaches approximately $346$ persons per frame while the average pedestrian density across the dataset is 178. A detailed sequence-wise summary of CroHD is given in Table \ref{tab:dataset_stats}. We split CroHD into 4 sequences of 5740 frames for training and 5 sequences of 5723 frames for testing. They share three scenes in common, while the fourth scene is disparate to ensure generalization of trackers on this dataset. A representative frame from each sequence of CroHD and their respective training, testing splits are depicted in Figure~\ref{fig:dataOverview}. We will make our sequences and training set annotations publicly available. To preserve the fairness of the MOTChallenge CVPR19 benchmark, we will not release the annotations corresponding to the test set. \\
\textbf{Annotation:} The annotation and data format of CroHD follows the standard guidelines outlined by MOTChallenge benchmark~\cite{MOT19_CVPR,MOT16}. We annotated all visible heads of humans in a scene with the visibility left to the best of discretion of annotators. Heads of all humans, whose shoulder is visible are annotated, including the heads occluded by head coverings such as hood, caps etc. For sequences inherited from MOTChallenge CVPR19 benchmark, the annotations were performed independent of pedestrian tracking ground truth in order to have no dependencies between the two modalities. Due to the high frame rate in our video sequences, we interpolate annotations in between keyframes and adjust a track only when necessary.\\
CroHD constitutes four classes - Pedestrian, Person on Vehicle, Static and Ignore. Heads of statues or human faces on clothing have been annotated with an ignore label. Heads of pedestrians on vehicles, wheelchairs or baby transport have been annotated as Person on Vehicle. Pedestrians who do not move throughout the sequence are classified as static persons. Unlike the case of standard MOTChallenge benchmarks, we observe that overlap between bounding boxes are minimal since head bounding boxes from an elevated viewpoint are almost distinct. Hence, we limit our visibility flag to be binary - either visible (1.0) or occluded (0.0). We consider a proposal to be a match if the Intersection Over Union (IoU) with the ground truth is larger than 0.4. 

\begin{table}
\tabcolsep=0.10cm
\begin{center}
\resizebox{\linewidth}{!}{%
 \begin{tabular}{|c c c c c c |} 
 \hline
Name & Frames & Scenario & Tracks & Boxes & Density\\ [0.5ex]
 \hline
CroHD-01 & 429 & Indoor & 85 & 21,456 & 50.0 \\

CroHD-02 & 3,315 & Outdoor, night & 1,276 & 733,622 & 222.0 \\

CroHD-03 & 1,000 & Outdoor, day & 811 & 258,012 & 258.0 \\

CroHD-04  & 997 & Indoor & 580 & 175,703 & 176.2  \\

CroHD-11 & 584 & Indoor & 133 & 38,492 & 65.8 \\

CroHD-12 & 2,080 & Outdoor, night & 737 & 383,677 & 185.0 \\

CroHD-13 & 1,000 & Outdoor, day & 725 & 257,828 & 258.0  \\

CroHD-14 & 1,050 & Outdoor, day & 562 & 258,227 & 246.0 
\\

CroHD-15 & 1,008 & Outdoor, day & 321 & 149,821 & 149.0  \\
\hline
Total  & 11,463 &  & 5,230 & 2,276,838 & 178    \\
\hline
\end{tabular}}
\end{center}
\caption{\label{tab:dataset_stats} Sequence-wise statistics CroHD. Sequences are named CroHD-XY, with X being either 0 or 1 depending on training set or testing set respectively. Y denotes the serial number of videos.}
\end{table}

\begin{figure}[htb]
\begin{center}
  \subfloat[Tracker A]{%
	   \centering
	   \includegraphics[width=0.48\linewidth]{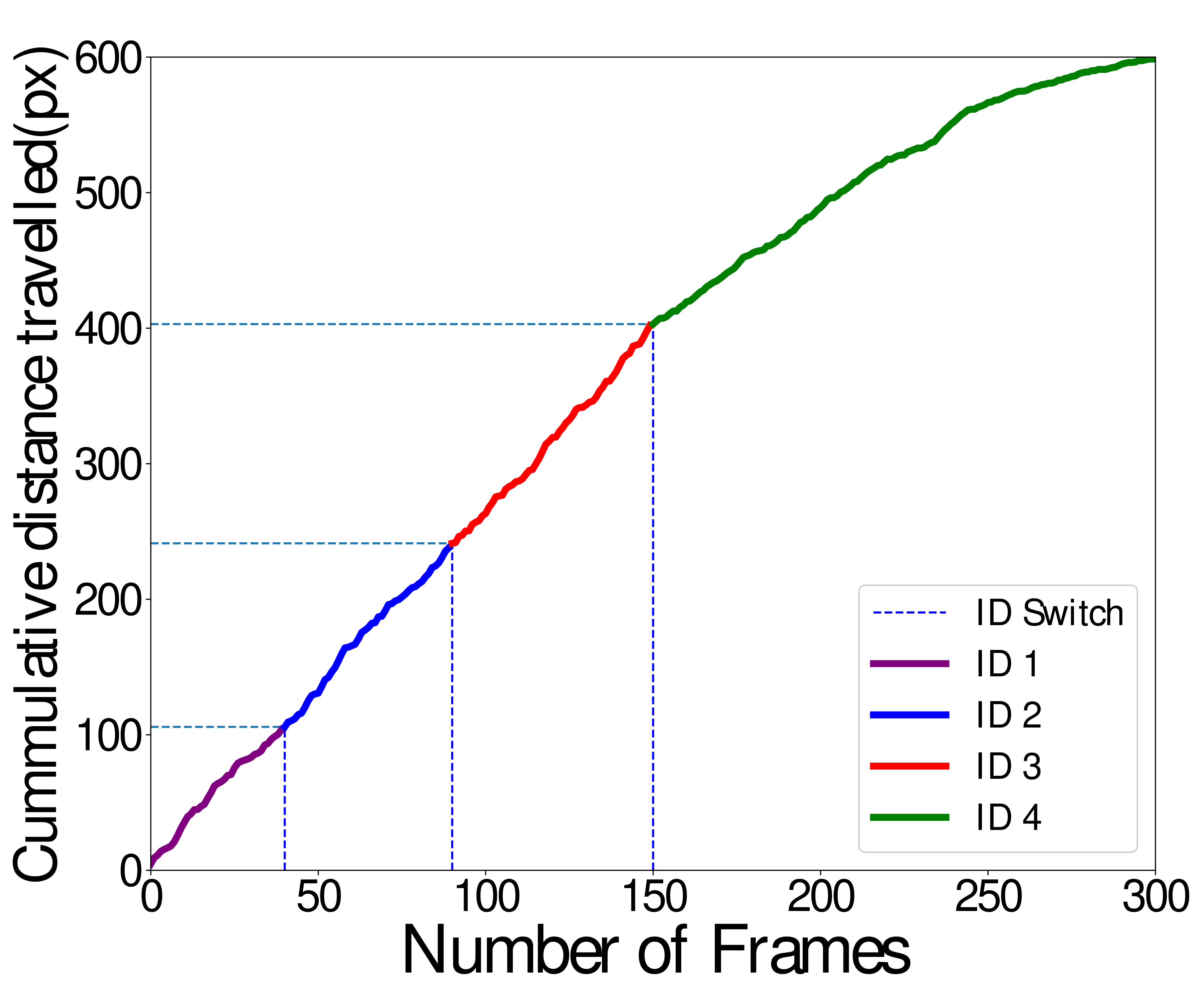}
	   \label{fig:gr1}}
 \hfill 	
  \subfloat[Tracker B]{%
	   \centering
	   \includegraphics[width=0.48\linewidth]{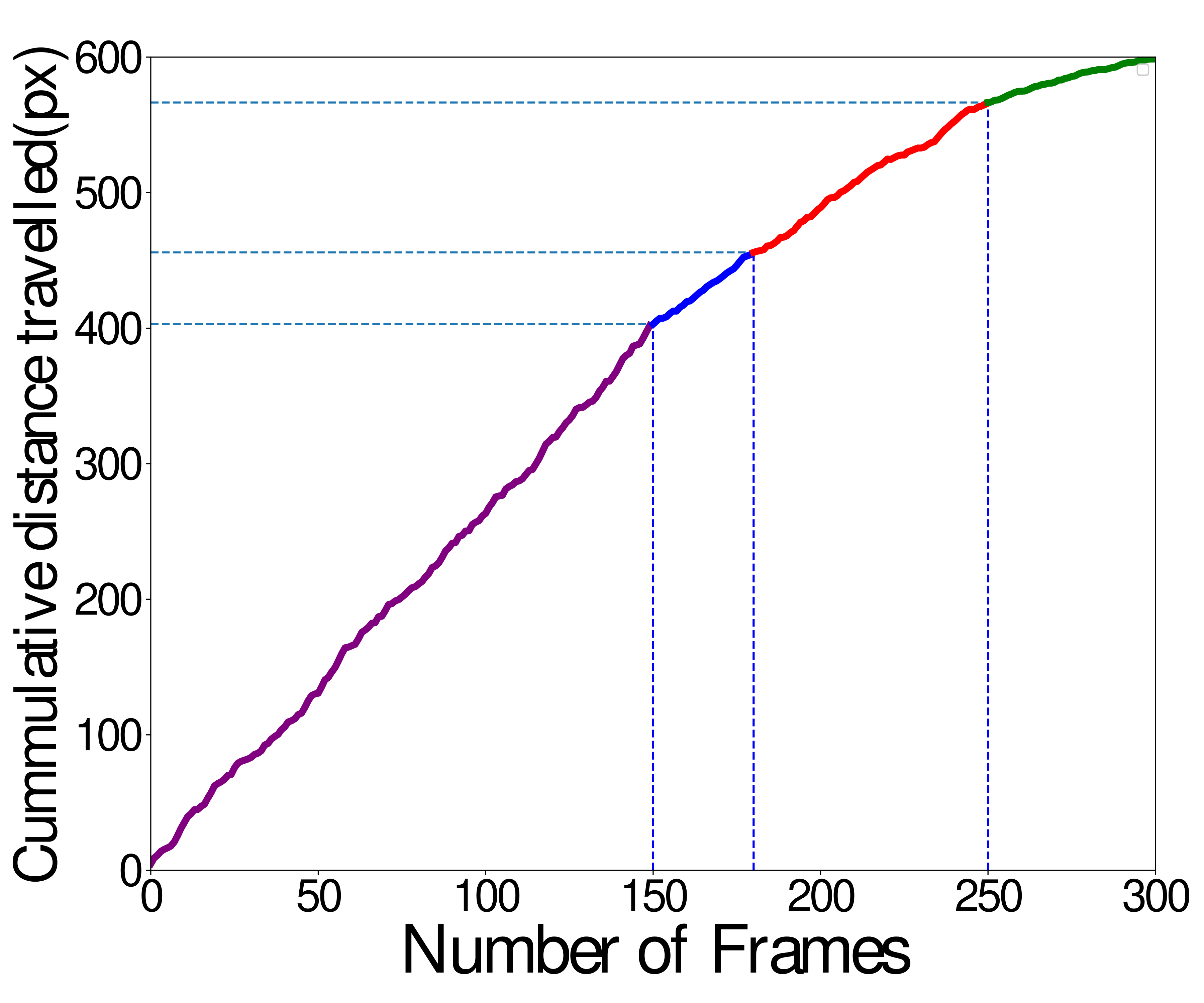}
	   \label{fig:gr2}}
\end{center}
\caption{Identity prediction of two trackers - Tracker A (\ref{fig:gr1}) and Tracker B (\ref{fig:gr2}) for the same ground truth. A change of color implies an identity switch with both trackers registering 3 switches.}
\label{fig:metric1}
\end{figure}

\section{Evaluation Metrics}
For evaluation of head detection on CroHD, we follow the standard Multiple Object detection metrics - mean Average Precision (mAP), Multiple Object Detection Accuracy (MODA), Multiple Object Detection Precision (MODP)~\cite{clearv5} and mAP\_COCO respectively. mAP\_COCO refers to a stricter metric which computes the mean of AP across IoU threshold of $\{50 \%, 55 \%, 60 \%, \ldots, 95 \%\}$. For evaluation of trackers, we adapt the well established Multiple Object Tracking metrics~\cite{CLEARMetric,DukeMTMC}, and extend with the proposed ``IDEucl'' metric.

\textbf{IDEucl:}
While the event based metrics~\cite{CLEARMetric} and identity based metric (IDF1)~\cite{DukeMTMC} are persuasive performance indicators of a tracking algorithm from a local and global perspective, they do not quantify the proportion of the ground truth trajectory a tracker in capable of covering. Specifically, existing metrics do not measure the proportion of ground truth trajectory in the image coordinate space a tracker is able to preserve an identity. It is important to quantitatively distinguish between trackers which are more effective in tracking a larger portion of ground truth pedestrian trajectories. This is particularly useful in dense crowds, for better understanding of global crowd motion pattern~\cite{crowd11_dataset}. To that end, we propose a new evaluation metric, ``IDEucl'', which gauges a tracker based on its efficiency in maintaining consistent identity over the length of ground truth trajectory in image coordinate space. Albeit, IDEucl might seem related to the existing IDF1 metric which measures the fraction of frames of a ground truth trajectory in which consistent ID is maintained. In contrast, IDEucl measures the fraction of the distance travelled for which the correct ID is assigned.

To elucidate this difference, consider the example shown in Figure~\ref{fig:metric1}. Two trackers A and B compute four different identities for a ground truth trajectory G. Tracker A commits three identity switches in the first 150 frames while maintaining consistent identity for the remaining 150 frames. Tracker B, on the other hand, maintains consistent identity for the first 150 frames but commits 3 identity switches in the latter 150 frames. Our metric reports a score of 0.3 for Tracker A (Figure~\ref{fig:gr1}) and a score of 0.67 for Tracker B (Figure~\ref{fig:gr2}). Meanwhile, IDF1 and the classical metric reports a score of ``0.5'' and ``3 identity switches'' respectively for both the trackers. Following existing metrics, Tracker A and Tracker B are considered equally efficient. They neither highlight the ineffectiveness of Tracker A nor the ability of Tracker B in covering an adequate portion of ground truth trajectory with consistent identity. Therefore, IDEucl is more appropriate for judging the quality of the estimated pedestrian motion.

Thus, to formulate this metric, we perform a global hypothesis to ground truth matching by constructing a Bipartite Graph $\mathcal{G} = (\mathcal{U}, \mathcal{V}, \mathcal{E})$, similar to~\cite{DukeMTMC}. Two ``regular" nodes are connected by an edge $e$ if they overlap in time, with the overlap defined by $\Delta$, 

\begin{equation}\Delta_{t, t-1} = \left\{\begin{array}{ll}
1, & \text { if } \delta > 0.5 \\
0, & \text { otherwise }
\end{array}\right.\end{equation}

Considering $\tau_{t}, h_{t}$ to be an arbitrary ground truth and hypothesis track at time $t$, $\delta$ is defined as,

\begin{equation}\delta= \mathrm{IoU}(\tau_{t}, h_{t}) \quad 
\end{equation}

The cost on each edge $\mathcal{E} \in \mathbb{R}^{N}$ of this graph, $\mathcal{M} \in \mathbb{R}^{N-1}$ is represented as the distance in image space between two successive temporal associations of ``regular'' node. In particular, cost of an edge is defined as , 

\begin{equation} \mathcal{M} = \sum_{t=1}^{N} m_{t} = \left\{\begin{array}{ll}
\mathrm{d}(\tau_{t}, \tau_{t-1}), & \text { if } \Delta_{t, t-1} = 1. \\
0, & \text { otherwise }
\end{array}\right.\end{equation}
where $\mathrm{d}$ denotes the Euclidean distance in image coordinate space. A ground truth trajectory is assigned a unique hypothesis which maintains a consistent identity for the predominant distance of ground truth in image coordinate space. We employ the Hungarian algorithm to solve this maximum weight matching problem to obtain the best (longest) hypothesis. Once we obtain an optimal hypothesis, we formulate the metric $\mathcal{C}$ as the ratio of length of ground truth in image coordinates covered by the best hypothesis,

\begin{equation}\mathcal{C} = \frac{\sum_{i=1}^{K} \mathcal{M}_{i}}{\sum_{i=1}^{K} \mathcal{T}_{i}} \end{equation}

Note that this formulation of cost function naturally weighs the significance of each ground truth track based on its distance in image coordinate space.

\section{Methods : Head Detection and Tracking}

In this section, we elucidate the design and working of HeadHunter and HeadHunter-T.

\subsection{HeadHunter}
\begin{figure*}[htb]%
\begin{center}
    \includegraphics[width=\textwidth]{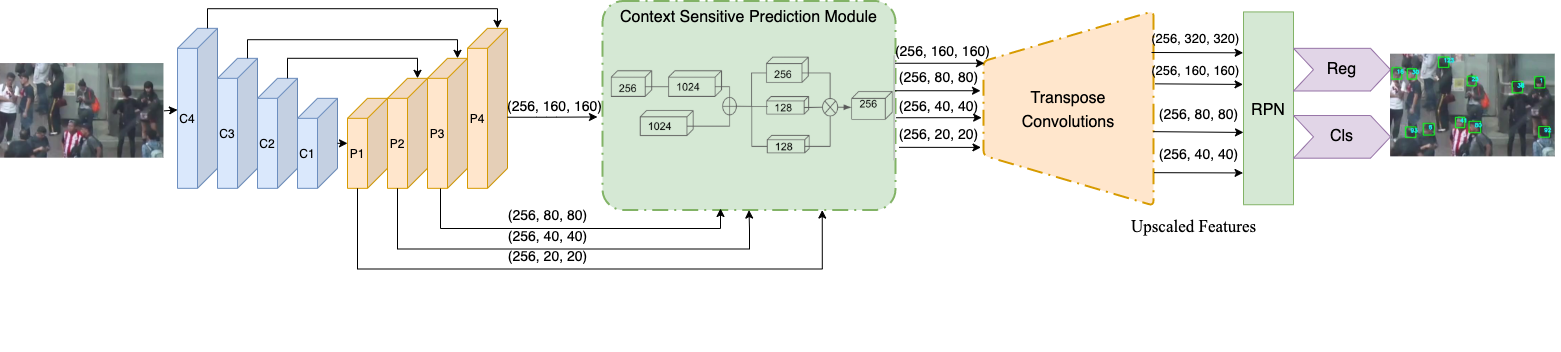}
\end{center}
    \caption{An overview of the architecture of our proposed head detector, Headhunter. We augment the features extracted using FPN (C4\dots P4) with Context Sensitive feature extractor followed by series of transpose convolutions to enhance spatial resolution of feature maps. Cls and Reg denote the Classification and Regression branches of Faster-RCNN~\cite{FRCNN} respectively.}
    \label{fig:our_architecture}
\end{figure*}

As detection is the pivotal step in object tracking, we designed HeadHunter differently from traditional object detectors~\cite{DPM,FRCNN,SDP} by taking into account the nature and size of objects we detect. HeadHunter is an end-to-end two stage detector, with three functional characteristics. First, it extracts feature at multiple scales using Feature Pyramid Network (FPN)~\cite{FPN} using a Resnet-50~\cite{ResNet} backbone. Images of heads are homogeneous in appearance and often, in crowded scenes, resemble extraneous objects (typically background). For that reason, inspired by the head detection literature, we augmented on top of each individual FPNs, a Context-sensitive Prediction Module (CPM)~\cite{PyramidBox}. This contextual module consists of 4 Inception-ResNet-A blocks~\cite{inception_resnet} with 128 and 256 filters for $3\times3$ convolution and 1024 filters for $1\times1$ convolution. As detecting pedestrian heads in crowded scenes is a problem of detecting many small-sized adjacently placed objects, we used Transpose Convolution on features across all pyramid levels to upscale the spatial resolution of each feature map. Finally, we used a Faster-RCNN head with Region Proposal Network (RPN) generating object proposals while the regression and classification head, each providing location offsets and confidence scores respectively. The architecture of our proposed network is summarised in Figure~\ref{fig:our_architecture}.

\subsection{HeadHunter-T}

We extended HeadHunter with two motion models and a color histogram based re-identification module for head-tracking. Our motion models consist of Particle Filter to predict motion of targets and Enhanced Correlation Coefficient Maximization~\cite{ECC} to compensate the Camera motion in the sequence. A Particle Filter is a Sequential Monte Carlo (SMC) process, which recursively estimates the state of dynamic systems. In our implementation, we represent the posterior density function by a set of bounding box proposals for each target, referred to as particles. The use of Particle Filter enables us to simultaneously model non-linearity in motion occurring due to rapid movements of heads and pedestrian displacement across frames.

\textbf{Notation: } 
Given a video sequence $\mathcal{I}$, we denote the ordered set of frames in it as $\{I_{0},\cdots, I_{T-1}\}$, where T is the total number of frames in the sequence. Throughout the paper, we use subscript notation to represent time instance in a video sequence. In a frame $I_{t}$ at time $t$, the active tracks are denoted by $\mathbf{T}_{t}=\{\mathbf{b}_{t}^{1}, \mathbf{b}_{t}^{2}, \ldots, \mathbf{b}_{t}^{N}\}$, where $\mathbf{b}_{t}^{k}$ refers to bounding box of the $k^{th}$ active track, denoted as $\mathbf{b}_{t}^{k}=\mathbf{\left(x_{t}^{k}, y_{t}^{k}, w_{t}^{k}, h_{t}^{k}\right)}$. At time $t$, the $i^{th}$ particle corresponding to $k^{th}$ track is denoted by $\mathbf{p}_{t}^{k,i}$ and its respective importance weight by $\mathbf{w}_{t}^{k,i}$. $\mathbf{L}_{t}$ and $\mathbf{N}_{t}$ denote the set of inactive tracks and newly initialized tracks respectively. 

\textbf{Particle Initialization: } New tracks are initialized at the start of the sequence, $I_{0}$ from the detection provided by HeadHunter and at frame $I_{t}$ for detection(s) which cannot be associated with an existing track. A plausible association of new detection with existing track is resolved by Non-Maximal-Suppression (NMS). The importance weights of each particle are set to be equal at the time of initialisation. Each particles represent 4 dimensional state space, with the state of each targets modelled as $(\mathbf{x}_{c}, \mathbf{y}_{c}, \mathbf{w}, \mathbf{h}, \mathbf{\dot{x}}_{c}, \mathbf{\dot{y}}_{c}, \mathbf{\dot{w}}, \mathbf{\dot{h}})$, where, $(\mathbf{x}_{c}, \mathbf{y}_{c}, \mathbf{w}, \mathbf{h})$ denote the centroids, width and the height of bounding boxes.

\textbf{Prediction and Update: }
At time $t>0$, we perform RoI pooling on the current frame's feature map, $\mathbf{F}_{t}$, with the bounding box of particles corresponding to active tracks. Each particles' location in the current frame is then adjusted using the regression head of HeadHunter, given their location in the previous frame. The importance weights of each particle are set to their respective foreground classification score from the classification head of HeadHunter. Our prediction step is similar to the Tracktor~\cite{tracktor}, applied to particles instead of tracks. Given the new location and importance weight of each particle, estimated position of $k^{th}$ track is computed as weighted mean of the particles,

\begin{align}
    \mathbf{S}_{t}^{k} = \frac{1}{M} \sum_{i=1}^{M} \mathbf{w}_{t}^{k,i} \mathbf{p}_{t}^{k,i}
\end{align}

\textbf{Resampling:} Particle Filtering frameworks are known to suffer from degeneracy problems~\cite{PF_tutorial} and as a result we resample to replace particles of low importance weight. $M$ particles corresponding to $k^{th}$ track are re-sampled when the number of particles which meaningfully contributes to probability distribution of location of each head, $\hat{\mathbf{N}}_{\mathrm{eff}}^{k}$ exceeds a threshold, where, 
\begin{equation}
    \hat{\mathbf{N}}_{\mathrm{eff}}^{k}=\frac{1}{\sum_{i=1}^{M}(\mathbf{w}^{k,i})^{2}}
\end{equation}

\textbf{Cost Matching:}
\label{costmatching}
Tracks are set to inactive when scores of their estimated state $\mathbf{S}_{t}^{a}$ falls below a threshold, $\lambda_{nms}^{reg}$. Positions of such tracks are predicted following Constant Velocity Assumption (CVA) and their tracking is resumed if it has a convincing similarity with a newly detected track. The similarity, $\mathbf{C}$ is defined as

\begin{equation}
 \mathbf{C} = \alpha \cdot IoU(\mathbf{L}_{t}^{i}, \mathbf{N}_{t}^{j}) + \beta \cdot d^{1}(\mathbf{L}^{i}_{t}, \mathbf{N}^{j}_{t})
 \label{eq:color_reid}
\end{equation}

where $\mathbf{L}_{t}^{i}$ and $\mathbf{N}_{t}^{j}$ are the $i^{th}$ lost track and $j^{th}$ new track respectively. And, $d^{1}$ denotes the Bhattacharyya distance between the respective color histograms in the HSV space~\cite{numiaro_histogram}. Once tracks are re-identified, we re-initialize particles around its new position.

\section{Experiments}

\subsection{HeadHunter}

We first detail the experimental setup and analyse the performance of HeadHunter on two datasets - SCUT-HEAD~\cite{scut_head} and CroHD respectively. For the Faster-RCNN head of HeadHunter, we used 8 anchors, whose sizes were obtained by performing K-means over ground truth bounding boxes from the training set. To avoid overlapping anchors, they were split equally across the four pyramid levels, with the stride of anchors given by $\max (16, s / d)$ where $s$ is square-root of the area of an anchor-box and $d$ is the scaling factor~\cite{fa-rpn}. For all experiments, we used Online Hard Example Mining~\cite{ohem} with 1000 proposals and a batch size of 512. \\

\textbf{SCUT-Head} is a large-scale head detection dataset consisting of 4405 images and 111,251 annotated heads split across Part A and Part B. We trained HeadHunter for 20 epochs with the input resolution to be the median image resolution of the training set (1000x600 pixels) and an initial learning rate of 0.01 halved at 5th, 10th and 15th epochs respectively. For a fair comparison, we trained HeadHunter only on the training set of this dataset and do not use any pre-trained models. We summarize the quantitative comparisons with other head detectors on this dataset in Table \ref{tab:headdetcmp_ScutHead}. HeadHunter outperforms other state-of-the-art head detectors based on Precision, Recall and F1 scores.

\begin{table}[htb]
\tabcolsep=0.45cm
\begin{center}
\resizebox{\linewidth}{!}{%
\begin{tabular}{lccc}
\hline 
Methods & Precision & Recall & F1 \\
\hline 
Faster-RCNN~\cite{FRCNN} & 0.87 & 0.80 & 0.83 \\


R-FCN+FRN~\cite{scut_head} & 0.91 & 0.84 & 0.87 \\

SMD~\cite{Sun_SMD} & 0.93 & 0.90 & 0.91 \\

HSFA2Net~\cite{HybridAttentionSelection} & 0.94 & 0.92 & 0.93 \\
\hline
\textbf{HeadHunter (Ours)} & \textbf{0.95} & \textbf{0.93} & \textbf{0.94} \\
\hline
\end{tabular}}
\end{center}
\caption{\label{tab:headdetcmp_ScutHead} Comparison between HeadHunter's and other state-of-the-art head detectors on the SCUT-Head dataset.}
\end{table}

\textbf{CroHD:} 
We first trained HeadHunter on the combination of training set images from SCUT-HEAD dataset and CrowdHuman dataset~\cite{shao2018crowdhuman} for 20 epochs at a learning rate of $0.001$. With variations well characterized, pre-training on large-scale image dataset improves the robustness of head detection. We then fine-tuned HeadHunter on the training set of CroHD, for a total of 25 epochs with an initial learning rate of $0.0001$ using the ADAM optimizer~\cite{ADAM_opt}. The learning rate is then decreased by a factor of $0.1$ at $10$th and $20$th epochs respectively. \\

\textbf{Ablation:} We examined our design choices for HeadHunter, namely the use of context module and the anchor selection strategy by removing them. The head detection performance of HeadHunter and its variants on CroHD are summarised in Table \ref{tab:baselinecmp_ourdata}. We threshold the minimum confidence of detection to 0.5 for evaluation. W/O Cont refers to the HeadHunter without Context Module. We further removed the median anchor sampling strategy and refer to as W/O Cont, mAn. We also provide baseline performance of Faster-RCNN with Resnet-50 backbone on CroHD, the object detector upon which we built HeadHunter. We followed the same training strategy for Faster-RCNN as HeadHunter. All variants of HeadHunter significantly outperformed Faster-RCNN. Inclusion of the context module and the anchor initialisation strategy also has a noteworthy impact on head detection.

\begin{table}[htb]
\tabcolsep=0.07cm
\begin{center}
\resizebox{\linewidth}{!}{%
\begin{tabular}{l c c c c c c}
\hline
Method & Precision & Recall & F1 & MODA & MODP & mAP\_COCO \\ 
[0.5ex]
\hline
 Faster-RCNN~\cite{FRCNN} & 34.4 & 42.2 & 50.1 & 40.3 & 30.8 & 11.2 \\
W/O Cont, mAn & 40.9 & 50.8 & 57.8 & 38.1 & 37.8 & 14.4\\
W/O Cont & 44.3 & 57.8 & 64.5 & 40.0 & 42.7 & 15.0\\
\hline
\textbf{HeadHunter} & \textbf{52.8} & \textbf{63.4} & \textbf{68.3} & \textbf{50.0} & \textbf{47.0} & \textbf{19.7} \\
\hline
\end{tabular}}
\end{center}
\caption{\label{tab:baselinecmp_ourdata} Summary of various head detector's performances on the test set of CroHD.}
\end{table}

\subsection{HeadHunter-T}

For the Particle Filtering framework, we used a maximum of N=100 particles for each object. The N particles were uniformly placed around the initial bounding box. To ensure that particles were not spread immoderately and were distinct enough, we sampled particles from a Uniform distribution whose lower and upper limit were $((x-1.5w, y-1.5h), (x+1.5w, y+-1.5h))$ respectively. Where, $x,y,w,h$ denote the centroid, width and height of the initial bounding box. For the color based re-identification, we used 16, 16 and 8 bins for the H, S and V channels respectively, where the brightness invariant Hue~\cite{invariant_hue} was used instead of the standard Hue. $\alpha$, $\beta$, which denotes the importance of IoU and color histogram matching, corresponding to Equation \ref{eq:color_reid} were set to 0.8 and 0.2 respectively. We deactivated a track if it remained inactive for $\lambda^{age}=25$ frames or if its motion prediction falls outside the image coordinates. \

We evaluated three state-of-the-art trackers on CroHD, namely, SORT~\cite{SORT}, V-IOU~\cite{v_iou} and Tracktor~\cite{tracktor} to compare with HeadHunter-T. We chose methods which do not require any tracking specific training, whose implementations have been made publicly available and are top-performing on the crowded MOTChallenge CVPR19 benchmark~\cite{MOT19_CVPR}. For a fair comparison, we performed all experiments with head detection provided by HeadHunter, thresholded to a minimum confidence of 0.6. SORT is an online tracker, which uses a Kalman Filter motion model and temporally associates detection based on IoU matching and Hungarian Algorithm. V\_IOU associates detection based on IoU matching and employs visual information to reduce tracking inconsistencies due to missing detection. Parameters for V\_IOU and SORT were set based on fine-tuning on the training set of CroHD, as discussed in the supplementary material. We evaluated two variants of Tracktor, with and without motion model. Tracktor+MM denotes the Tracktor extended with Camera Motion Compensation~\cite{ECC} and CVA for inactive tracks. For the two versions of Tracktor, we set tracking parameters similar to HeadHunter. Table \ref{tab:baseline_ours} summarises the performance of aforementioned methods on the test set of CroHD. HeadHunter-T outperforms all the other methods, and furthermore demonstrates superiority in identity preserved tracking. Although Tracktor~\cite{tracktor} is similar to HeadHunter-T, there is a noticeable difference in its head tracking performance. We hypothesize the use of Particle Filter framework, which can handle arbitrary posteriors, as the reason for improvement. This claim is justified in the forthcoming section.

\begin{table}
	\begin{center}
	\tabcolsep=0.05cm
	\resizebox{\columnwidth}{!}{%
		\label{tab:baseline_ours}
		\begin{tabular}{l c c c c c c }
		\hline
		 Method & MOTA $\uparrow$ & IDEucl $\uparrow$ & IDF1 $\uparrow$ & MT $\uparrow$ & ML $\downarrow$ & ID Sw. $\downarrow$ \\ [0.5ex]
			\hline
			SORT~\cite{SORT} & 46.4 & 58.0 & 48.4 & 49 & 216 & \textbf{649} \\ [1pt]

			V\_IOU~\cite{v_iou} & 53.4 & 34.3 & 35.4 & 80 & 182 & 1890 \\

			Tracktor~\cite{tracktor} & 58.9 & 31.8 & 38.5 & 125 & 117 & 3474 \\

			Tracktor+MM~\cite{tracktor} & 61.7 & 44.2 & 45.0 & 141 & 104 & 2186 \\
			\hline
			\textbf{HeadHunter-T} & \textbf{63.6} & \textbf{60.3} & \textbf{57.1} & \textbf{146} & \textbf{93} & 892 \\ \hline
	\end{tabular}}
	\end{center}
	\caption{\textbf{Main tracking result} comparing the performances of various state-of-the-art trackers and HeadHunter-T on the test set of CroHD. The direction of arrows indicate smaller or larger desired value for the metric.}
\end{table}

\subsection{Ablation Experiments}

\textbf{HeadHunter-T:}
In this section, we analyse the design choices, in particular, the utility of re-identification module and Particle Filter of HeadHunter-T on the training set of CroHD. The results are summarised in Table \ref{tab:ablation_tracker}. For variations in motion model, we removed the Particle Filter and used simple Camera Motion Compensation, denoted as HT w/o PF. We also experimented with a reduced number of particles initialized around the head, with n=10, denoted as HT + 10F. Introducing Particle Filter noticeably improved identity preserving scores (IDF1 and IDEucl) for HT + 10F. Further increasing the number of filters to 100 demonstrated the best performance. However, using more than 100 filters resulted in either duplicates or immoderate spreading, which are undesirable. We removed the re-identification module, to understand its influence, denoted as w/o ReID. Although color histogram is a modest image descriptor, yet it drastically reduced the number of identity switches and showed superior performance in identity preserving metrics - IDEucl, IDF1. We also experimented with $\alpha$ and $\beta$ values corresponding to the importance of IoU and histogram matching (Equation~\ref{eq:color_reid}). We set $\beta$ to 0.8 and $\alpha$ to 0.2 and this configuration is denoted as HT + sReID. Surprisingly, we observed more identity switches and a slight decrease in performance across other tracking metrics. HeadHunter-T, our final model, outperformed all the other variants.

\begin{table}
\begin{center}
\tabcolsep=0.05cm

    \resizebox{\columnwidth}{!}{
    \begin{tabular}{l c c c c c c }
        \hline
        Method & MOTA $\uparrow$ & IDEucl $\uparrow$ & IDF1 $\uparrow$ & MT $\uparrow$ & ML $\downarrow$ & ID Sw. $\downarrow$ \\ [0.5ex] 
        \hline
        HT w/o PF  & 60.6 & 40.1 & 43.9 & 200 & 102 & 3652 \\ 

        HT + 10F  & 63.3 & 58.2 & 56.3 & 214 & 98 & 1534 \\
        
        \hline
        
        HT w/o ReID & \textbf{59.5} & 57.7 & 57.5 & \textbf{225} & \textbf{91} & 1411 \\

        HT + sReID & 59.1 & 57.8 & 58.3 & \textbf{225} & \textbf{91} & 1280 \\

        \hline

        HT + KF & 63.4 & 53.8 & 55.9 & 214 & 93 & 2451 \\ 
        \hline
        \textbf{HeadHunter-T} & \textbf{64.0} & \textbf{61.5} & \textbf{58.5} & \textbf{225} & \textbf{91} & \textbf{1247} \\
        \hline
    \end{tabular}}
\end{center} 
\caption{Illustration of ablation studies of HeadHunter-T (denoted as HT) on the training set of CroHD. The direction of arrows indicate small or large desired metric values.}

\label{tab:ablation_tracker}
\end{table}

\textbf{Choice of Filter: }
To further substantiate our choice of a multi-modal filter, we replaced the Particle Filter of HeadHunter-T with a Kalman Filter motion model~\cite{KalmanFilter}. While both Kalman Filter and Particle Filter are recursive state estimation algorithms, Kalman Filter assumes the system to be linear with Gaussian noise~\cite{PF_tutorial} while Particle Filter's multimodal posterior distribution enables it to model states of nonlinear systems. We replaced the Particle Filter with a four state Kalman Filter to model the inter-frame displacement of bounding boxes with CVA. The four states are $x,y$ centroid coordinates, the height and aspect ratio of bounding boxes respectively, similar to the SORT~\cite{SORT}. The performance of this tracker, denoted as HT + KF is summarised in Table \ref{tab:ablation_tracker}. HeadHunter-T with Particle Filter demonstrates superior performance than its Kalman Filter variant with respect to all the tracking metrics reported and in particular, we observe major improvement in-terms of IDEucl metric. Motion of heads along with the pedestrian displacement induces non-linearities in the position of bounding boxes. Although pedestrian motion in general is non-linear, this issue is exacerbated with the small size of head bounding boxes. Hence, using a multi-modal posterior state estimation is necessary to address the perceptible impact of non-linear motion. We remark this to be the reason behind improvement in performance while using a Particle Filter in comparison to the Kalman Filter.
\\

\begin{figure}[htb]
\begin{center}
\includegraphics[width=\linewidth]{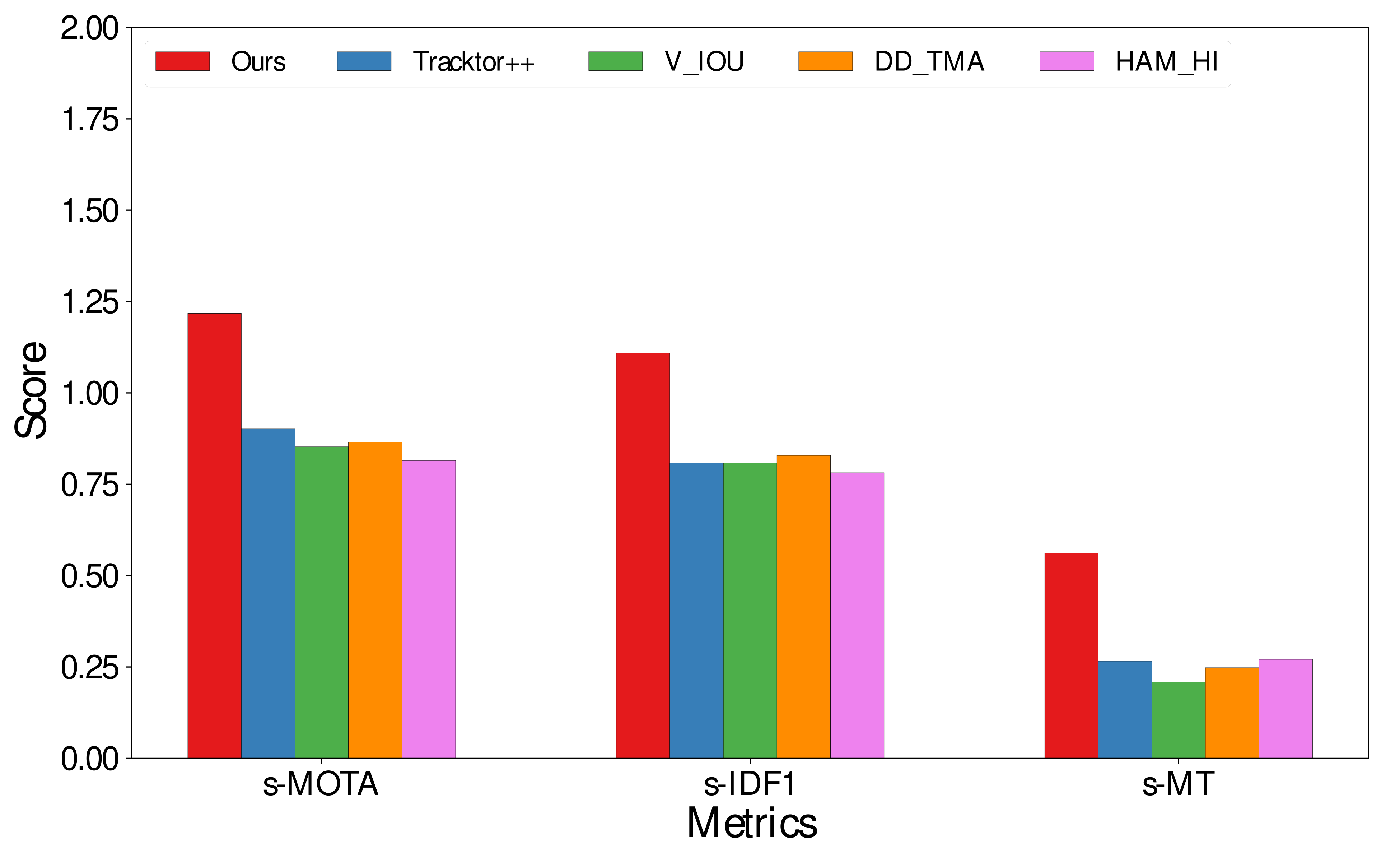}
\end{center}
\caption{Comparison between HeadHunter-T and state-of-the-art trackers on common sequences of CroHD and MOTChallenge benchmark~\cite{MOT19_CVPR}. s-MOTA, s-IDF1, s-MT are scaled version of MOTA, IDF1 and Most Tracked (MT) metrics respectively.}
\label{fig:tracker_compare}
\end{figure}

\textbf{Comparison across paradigm :}
We compare pedestrian and head tracking performances on the common sequences between CroHD and MOTChallenge CVPR19 dataset. The sequence being the same ensures that trackers are evaluated on full body and head bounding boxes of the same pedestrians in the scene. For this comparison, we chose published state-of-the-art methods on the aforementioned dataset, namely, Tracktor++~\cite{tracktor}, V\_IOU, DD\_TMA~\cite{dd_tama} and HAM\_HI~\cite{HAM_HI}. We performed comparison in-terms of MOTA, IDF1, MT (Mostly Tracked in percentage) metrics. Since we used a different object detector than the rest, a straightforward comparison between performance metrics would not be fair. Hence, for each sequence, we measure the ratio of aforementioned performance metrics with their object detector's MODA score to obtain the scaled scores - s-MOTA, s-IDF1 and s-MT. The scaled scores, averaged across five common sequences are illustrated in Figure~\ref{fig:tracker_compare}. Our approach substantially outperforms other methods indicating that tracking by head detection is more suited for tracking in environments involving high pedestrian density where preserving identity is important. It is also worthy to note that HeadHunter uses a ResNet-50 backbone in contrast to a Resnet-101 backbone used by other methods. Furthermore, Tracktor++, HAM\_HI and DD\_TMA all use Deep Networks for extracting appearance features, while HeadHunter-T uses a color histogram based appearance feature. By compromising our tracking space (size of bounding box) to avoid mutual occlusion, we observe notable performance gain and significantly reduce the computation cost. This suggests that tracking by head detection paradigm is more desirable for real-time tracking applications focused on identity preservation.

\section{Conclusion}

To advance algorithms to track pedestrians in dense crowds, we introduced a new dataset, CroHD, for tracking by head detection. To further quantify the efficacy of a tracker in describing pedestrian motion, we introduced a new metric, IDEucl. We developed two new baseline methods, HeadHunter, HeadHunter-T for head detection and head tracking on CroHD respectively. We demonstrated HeadHunter-T to be consistently more reliable for identity preserving tracking applications than existing state-of-the-art trackers adapted for head tracking. Additionally, the adequacy of HeadHunter-T's performance with a modest computational complexity, opens up opportunities for future research focused on tracking methods adapted for low computational complexity and real-time applications. We also hope that CroHD will serve useful in contiguous fields, such as Crowd Counting and Crowd Motion Analysis.

{\footnotesize \paragraph{Acknowledgements:} We are thankful to Dr. Vicky Kalogeiton and Prof. Dr. Bastian Leibe for their insightful feedback. We are also thankful to our annotators for their hard work. This work has received funding from the European Union’s Horizon 2020 research and innovation programme under the grant agreement No. 899739.}

\part*{Supplemental Material}
\input{supplementary.tex}
\end{document}

%% file: supplementary.tex
\maketitle

\begin{abstract}
In this supplementary material, we provide more detailed insights into the statistics of our dataset and its annotation procedure. We also report the influence of hyperparameters of trackers, which we have used for performing baseline experiments. Finally, we explore the role of various head detectors in tracking performances and present the sequence-wise result of HeadHunter and HeadHunter-T on CroHD.

\end{abstract}

\section{CroHD Annotation}
\begin{figure*}[htb]
\begin{center}
\begin{adjustbox}{minipage=\linewidth,scale=0.95}
\begin{minipage}{.49\linewidth}
\centering
\includegraphics[width=.98\linewidth]{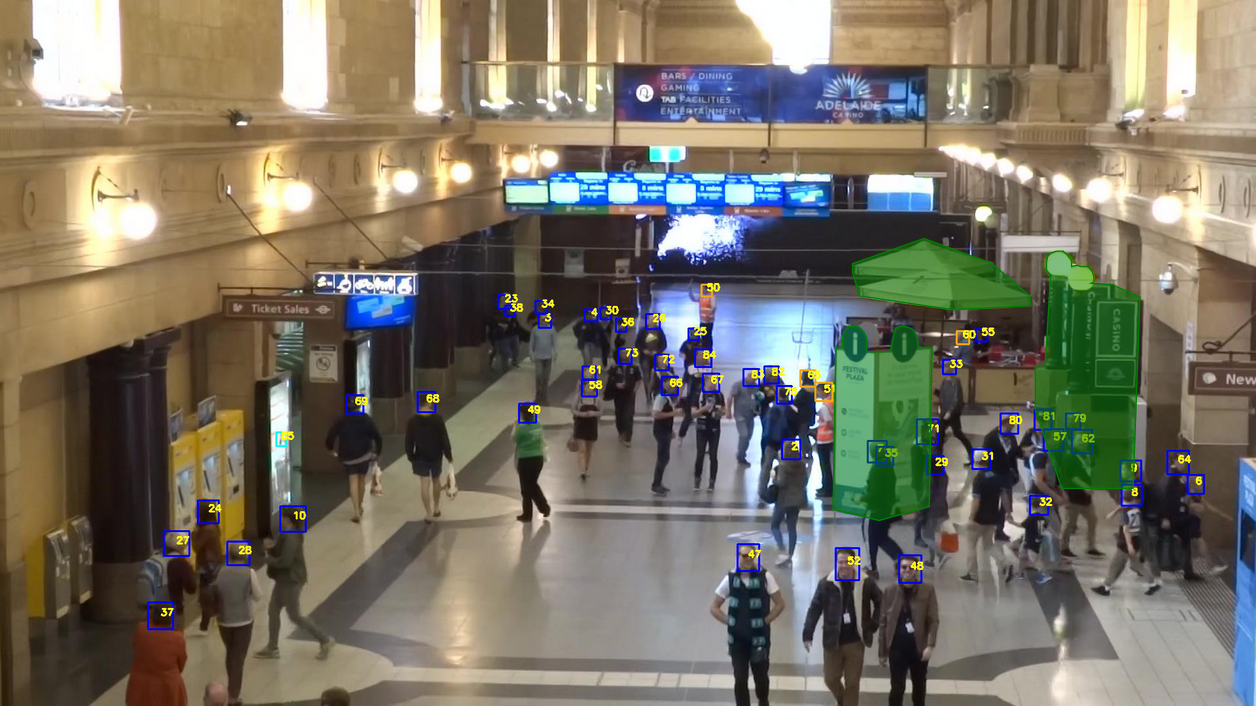}
\end{minipage}%
\begin{minipage}{.49\linewidth}
\centering
\includegraphics[width=.98\linewidth]{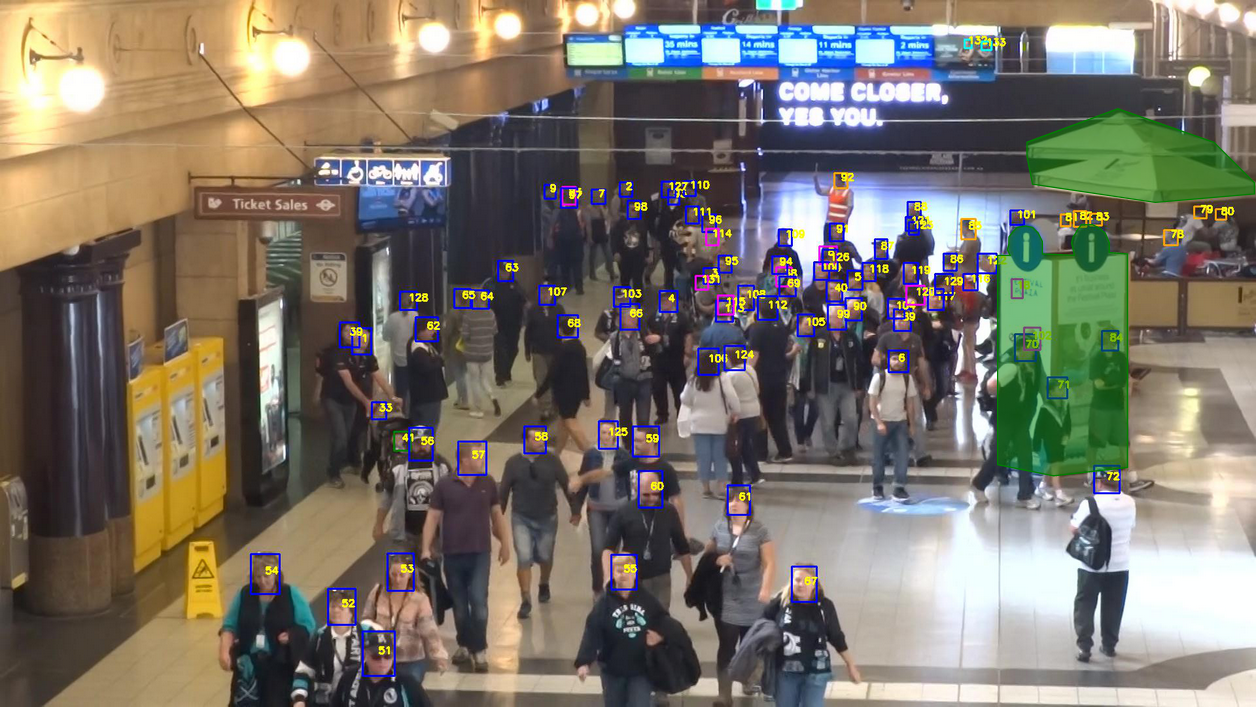}
\end{minipage}\par\medskip
\begin{minipage}{.49\linewidth}
\centering
\includegraphics[width=.98\linewidth]{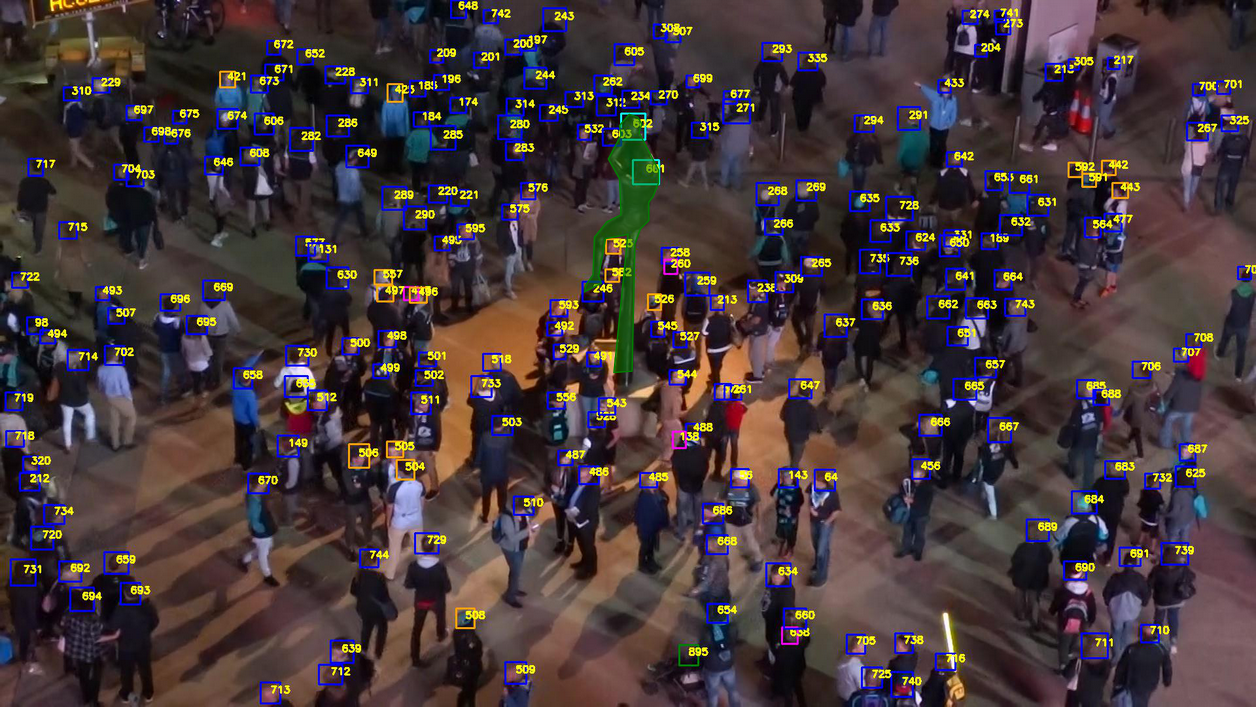}
\end{minipage}%
\begin{minipage}{.49\linewidth}
\centering
\includegraphics[width=.98\linewidth]{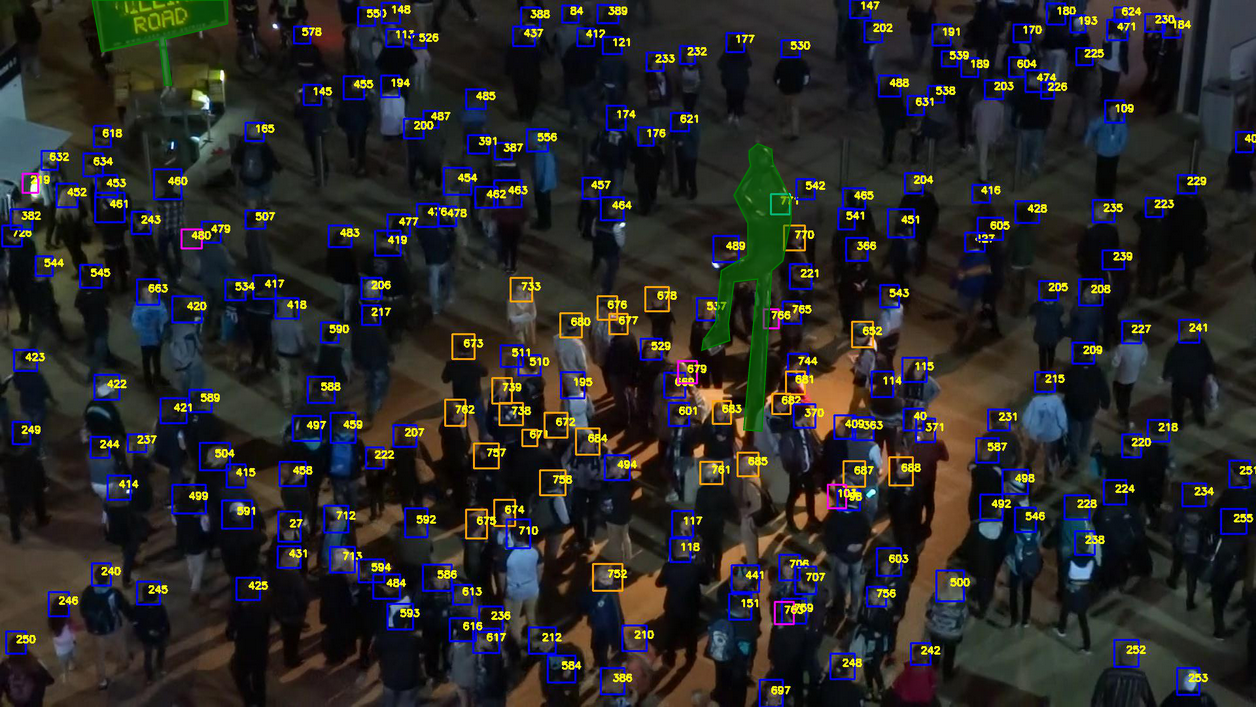}
\end{minipage}\par\medskip
\begin{minipage}{.49\linewidth}
\centering
\includegraphics[width=.98\linewidth]{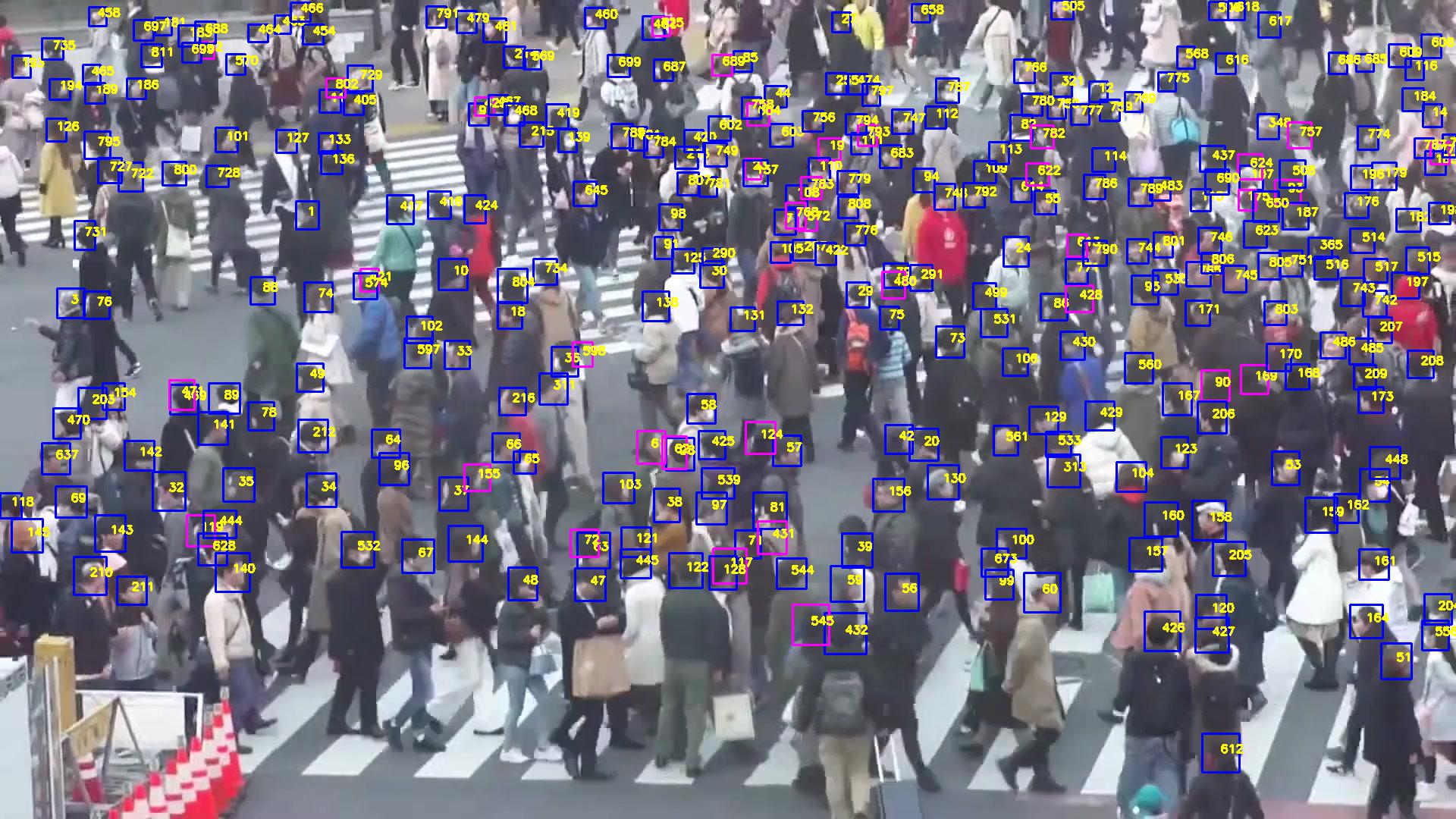}
\end{minipage}%
\begin{minipage}{.49\linewidth}
\centering
\includegraphics[width=.98\linewidth]{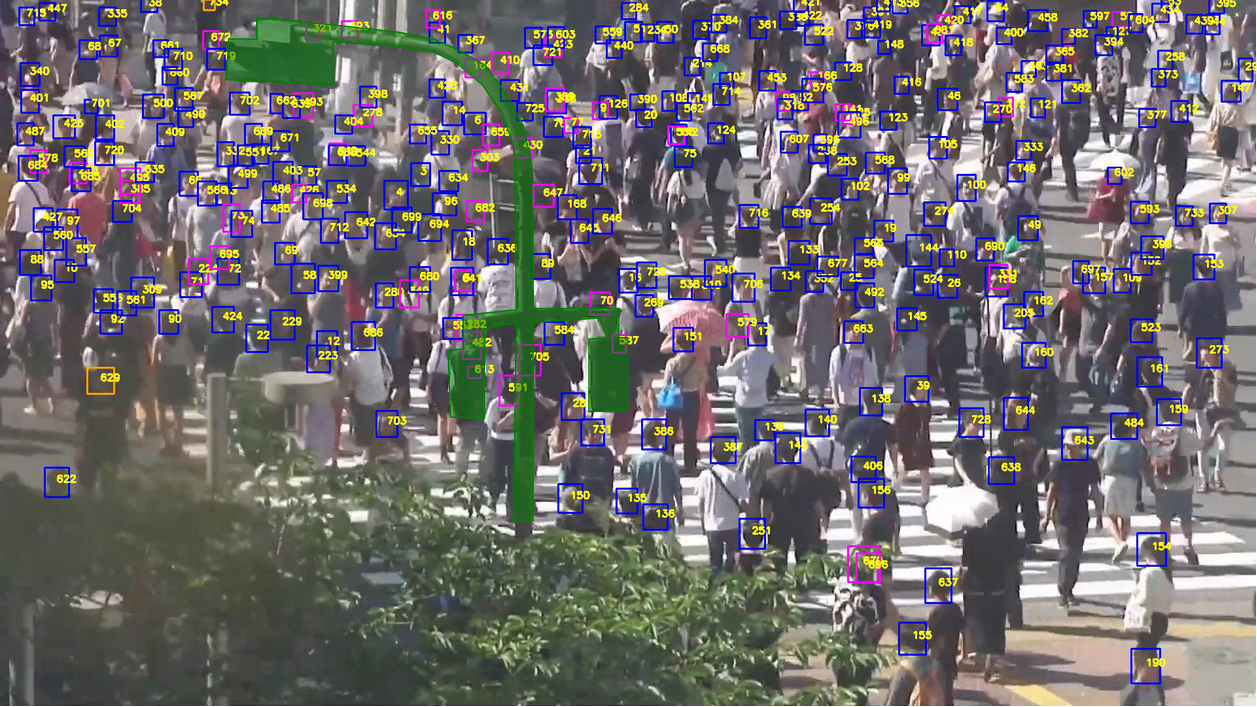}
\end{minipage}\par\medskip
\begin{minipage}{.49\linewidth}
\centering
\includegraphics[width=.98\linewidth]{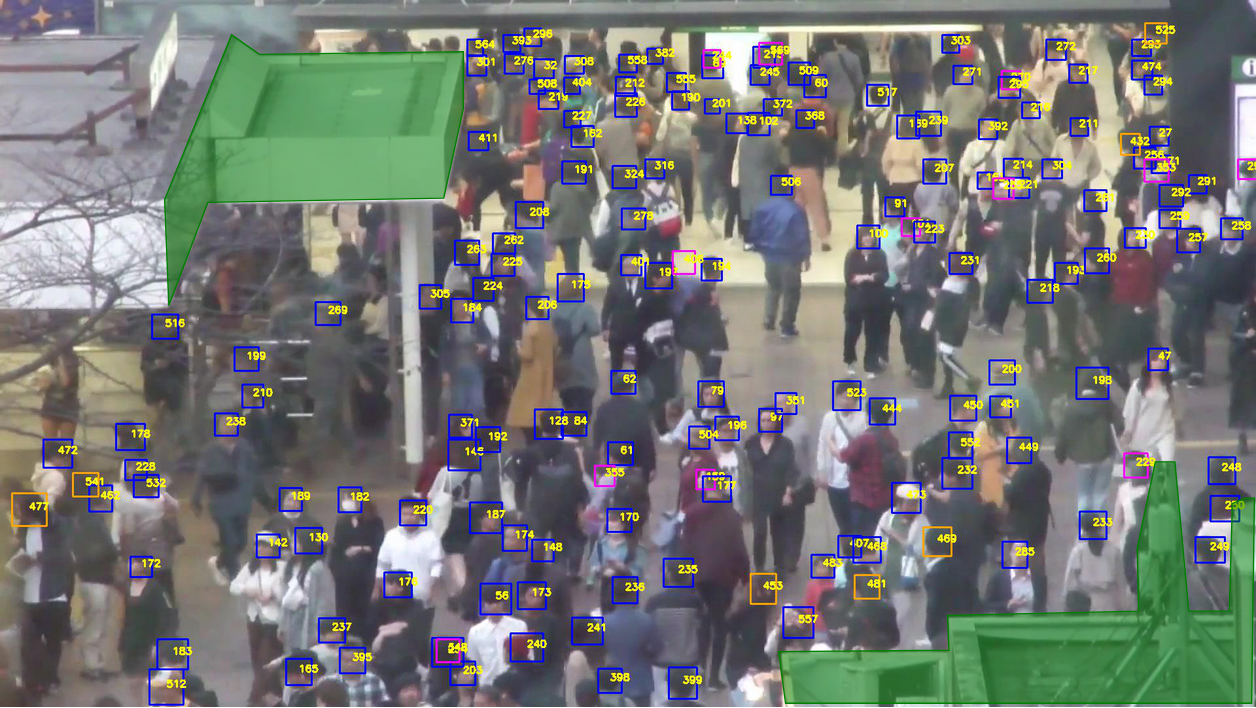}
\end{minipage}%
\begin{minipage}{.49\linewidth}
\centering
\includegraphics[width=.98\linewidth]{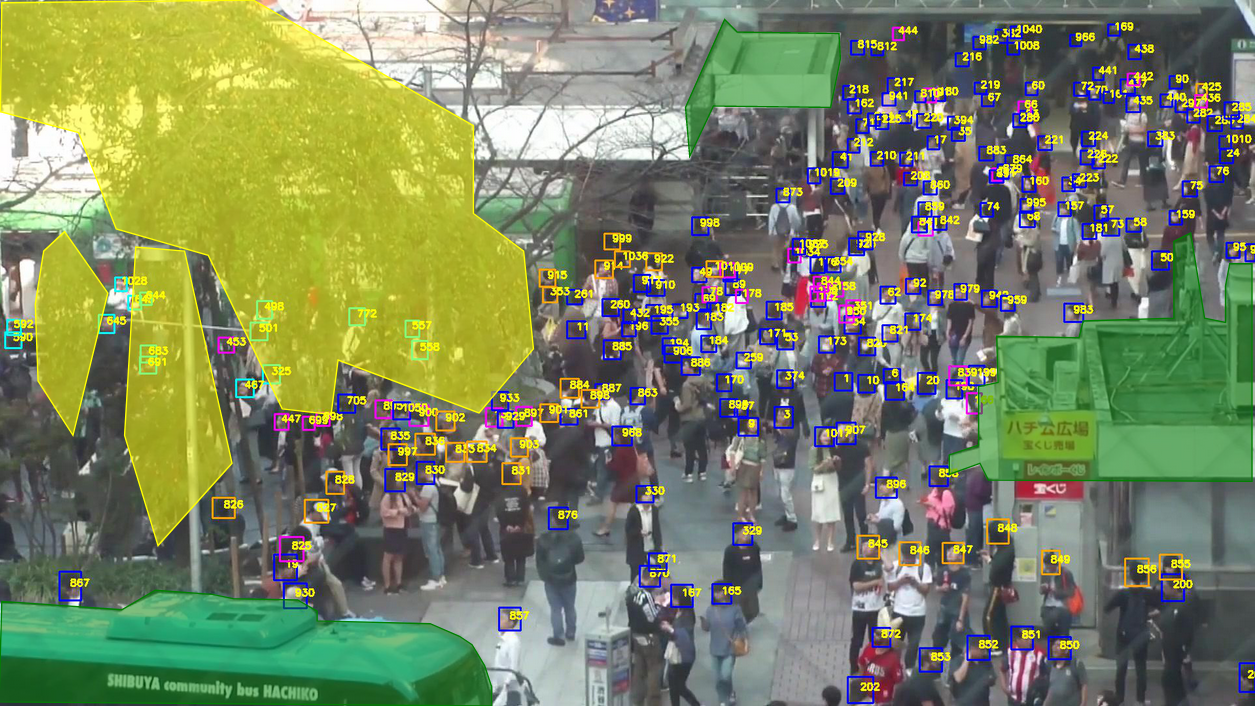}
\end{minipage}\par\medskip
\begin{minipage}{.49\linewidth}
\centering
\includegraphics[width=.1\linewidth]{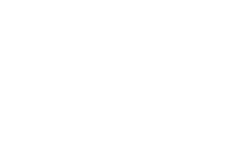}
\end{minipage}
\begin{minipage}{.49\linewidth}
\centering
\includegraphics[width=.98\linewidth]{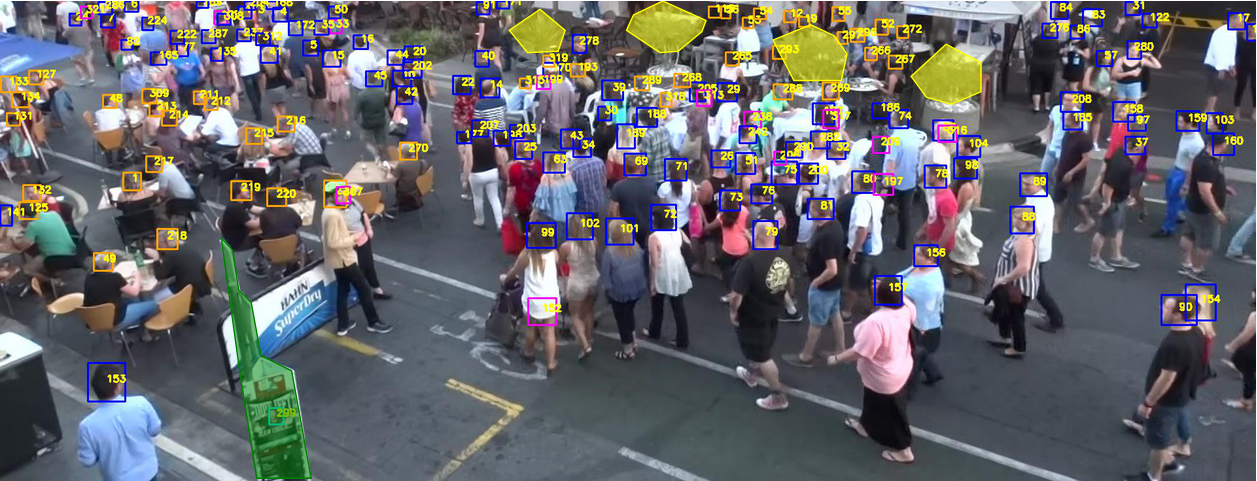}
\end{minipage}\par\medskip
\end{adjustbox}
\end{center}
\caption{An overview of annotated frames from our dataset, CroHD. In both train (left column) and test (right column) sets, bounding boxes of heads are either active (dark blue), static (orange), occluded (pink) or non-human (light blue). Occluders are present in many scenes, either opaque (green) or translucent (yellow). }

\label{fig:occlusion_region}
\end{figure*}

We annotated heads of pedestrians in this dataset in order to reduce the intra-target occlusions. The annotation work was performed with the help of Crowdsourcing platform, Fiverr \footnote{\url{http://fiverr.com/}} using the CVAT Annotation tool \footnote{\url{https://github.com/opencv/cvat}}. Due to the number of targets to be tracked being plentiful, while the area of tracking is significantly smaller than existing approaches, the margin for errors in this annotation procedure is large. As a result, we employed a three-stage reviewing process for thoroughgoing the annotation. First, we automated the process of spotting identity switches and track fragmentations, which were the most common mistakes made by annotators. Then, the annotations corresponding to a sequence were reviewed by a team of annotators, separate from those who annotated the particular scene, to avoid any bias. Finally, we (the authors of this work) manually inspected the annotation.\\
\textbf{Automation of reviewing: } A pedestrian head is assigned an ID as soon as it becomes visible and the same ID is maintained until it leaves the field of view (FoV). Using this information, we gathered tracks which have not terminated near the image boundary, with the last few frames being an exception. This helped us in identifying tracks whose annotations have been fragmented. Another common mistake in annotations were identity switches, when the identity of two pedestrian heads end up mutually swapping. In order to spot this, for each target, we analyzed the displacement of respective bounding box centroids. If at a particular frame, the motion of a particular track was two standard deviations away from the mean displacement, such tracks were flagged for a potential identity switch review. Note that both methods mentioned in this section are not complete and do not recognize all fragmentation and identity switches. However, they have significantly helped in minimizing human efforts in spotting such errors. \\
\textbf{Visibility:} Figure \ref{fig:occlusion_region} shows an example of various types of occluders across all scenes in our dataset. Occluders in the scene, which are either opaque or translucent, affect the visibility of pedestrians. Heads obscured by Translucent occluders such as tree leaves were annotated with the ``ignore label'' for tracking but are considered for evaluation of head detectors. Heads obscured by opaque occluders were neither considered for the evaluation of tracking nor detection and are annotated with visibility flag of ``0''. Assigning a visibility flag for a heads was left to the best discretion of annotators. \\
\textbf{Key Frame Annotation:}
Due to the high frame rates (25 FPS) across videos, we employ keyframe annotation rule, with every $10^{th}$ frame considered a keyframe. Annotations were performed only on keyframes with a linear interpolation employed to annotate the positions of bounding boxes for the frames in between two successive keyframes. We used every $5^{th}$ frame to be a keyframe in sequences CroHD-03 and CroHD-13, where the pedestrian density and velocity are significantly higher than the other sequences, and parts of sequences where minor camera motion was incurred. Bounding boxes were adjusted in between keyframes for pedestrians in a particular frame if needed due to perceptible head motion. Once annotations were completed for a particular scene, two separate annotators reviewed the frames in between keyframes to supervise termination, initialization and occlusion handling of tracks. \\
\textbf{Statistics:}
We analyze the detailed statistics of our benchmark in this section as summarized in Table \ref{tab:benchmark_robust_stat}. Specifically we look into the statistics of our track length, pedestrian velocities, bounding box ratio, occlusions and class distribution. Average pedestrian velocity is the mean distance travelled by the tracks between each frame in pixels, averaged over the whole sequence and represented as $px.s^{-1}$. Bounding box ratio (BBR) denotes the ratio of spatial dimensions of frames to that of average bounding box in the respective sequence. Occlusion refers to the average time (in frames) that a target was annotated with a visibility flag of ``0''. \\
We compare CroHD with multiple pedestrian tracking benchmarks based on number of pedestrian annotations, pedestrian densities and tracks annotated as depicted in Table~\ref{tab:tracking_benchmark_comparison}. The density in the table refers to the average number of pedestrian annotations per frame. CroHD has the largest pedestrian annotation, pedestrian density and number of tracks.

\begin{table}[htb]
\tabcolsep=0.15cm
\begin{center}
\resizebox{\linewidth}{!}{%
\begin{tabular}{l c c c c c c|}
\hline
Dataset & Videos & Frames & Boxes & Density & Tracks \\
\hline
MOTChallenge-15 \cite{MOTChallenge2015}& \textbf{22} & 11,283 & 101,345 & 8.95 & 1221 \\

MOTChallenge-16 \cite{MOT16}& 14 & 11,235 & 292,733 & 25.8 & 1342 \\

MOTChallenge-19\cite{MOT19_CVPR}& 9 & \textbf{13,410} & 2,259,143 & 171.0 & 3882\\

MOTS\cite{MOTS}& 8 & 5,906 & 59,163 & 10.0 & 578\\
\hline
\textbf{CroHD}& 9 & 11,463 & \textbf{2,276,838} & \textbf{178.0} & \textbf{5230} \\
\hline
\end{tabular}}
\end{center}
\caption{\label{tab:tracking_benchmark_comparison} Comparison between CroHD and existing multiple-pedestrian tracking benchmarks. Barring density, all the other columns refer to total figures for respective benchmarks.}
\end{table}

\section{Hyperparameter Tuning}
\label{sec:hyper_param}
In this section, we discuss the influence of hyperparameters for trackers which we used for baseline experiments on CroHD - IoU Tracker~\cite{v_iou} and SORT~\cite{SORT}. For the two experiments, we used the detection provided by HeadHunter, to ensure fairness in evaluation.

\subsection{IoU Tracker}

We mainly study the influence of parameter $\sigma_{iou}, \sigma_{h}, \mathrm{\textit{ttl}} $ and $t_{min}$. The minimum IoU between two detection overlaps to be considered a track is denoted by $\sigma_{iou}$. Tracks are filtered if they do not contain at least one detection with an IoU $\geq \sigma_{h}$ for at least $t_{min}$ frames. \textit{ttl} denotes the number of frames through which visual tracking is performed backwards, with the Kernelized Correlation Filters (KCF)~\cite{KCF} applied for visual tracking. We observe no noticeable change with modification of parameters $\sigma_{h}$ and \textit{ttl}. We further attempted MedianFlow~\cite{MedianFlow}, TLD~\cite{TLD} as choices for visual tracking and no significant changes were observed with these modifications either. We hypothesize the size of objects being tracked as a reason for the observed invariance in performances. The results are summarized in Table~\ref{tab:viou}. First row shows the performance of this tracker with all hyperparameters set to their default value. Better performance with respect to the identity metric are observed in the case of default $t_{min}$ value while a lower $t_{min}$ and higher $\sigma_{iou}$ signifies a better MOTA score.

\begin{table}[htb]
\tabcolsep=0.25cm
\begin{center}
\resizebox{0.8\linewidth}{!}{%
\begin{tabular}{c c c c c} 
\hline

$\sigma_{iou}$ & $t_{min}$ & MOTA & IDEucl & IDF1 \\ [0.5ex]
\hline
0.3  & 5 & 51.0 & 31.9 & 33.7\\
\hline
0.2 & \deemph{5} & 51.4 & \textbf{32.6} & \textbf{34.1}\\

0.4 & \deemph{5} & 50.1 & 28.8 & 32.2\\

0.5 & \deemph{5} & 48.0 & 23.6 & 29.0\\

0.8 & \deemph{5} & 42.5 & 17.1 & 23.6\\
\hline

\deemph{0.3} & 4 & 51.6 & 30.9 & 33.6\\

\deemph{0.3} & 3 & 52.1 & 30.2 & 33.4\\

\deemph{0.3} & 2 & \textbf{52.4} & 29.1 & 33.2\\
\hline
\end{tabular}}
\end{center}
\caption{\label{tab:viou} Results of tuning V\_IOU\cite{v_iou} tracker's hyper-parameters on the training set of CroHD.}
\end{table}

\subsection{SORT}

We analyze three parameters corresponding to SORT~\cite{SORT}, namely, max\_age, min\_hits and min\_IoU. The maximum age a track will be kept alive without being associated to a detection is denoted by max\_age. Without an associated detection, the position of tracks are updated through a Kalman Filter framework following Constant Velocity Assumption (CVA) for max\_age frames. The minimum IoU required between subsequent detection of a particular track is denoted by min\_IoU and min\_hits denotes the number of minimum subsequent detection required to be associated to initialize a track. Table~\ref{tab:sort} summarizes the performance of SORT with varying hyperparameters. The first row corresponds to the default configuration while the last row denotes the best amongst the configurations we have varied. A straightforward observation is improvement with increasing max\_age, more notably in-terms of IDEucl metrics. This is in contrast with what Bewely \etal~\cite{SORT} remark in their original paper. Furthermore, a significant improvement is also observed by reducing the min\_IoU. These two occurrences can be explained due to significantly reduced overlaps between bounding boxes in tracking by head detection paradigm compared to tracking by full-body detection. 

\begin{table}[htb]
\tabcolsep=0.15cm
\begin{center}
\resizebox{\linewidth}{!}{%
\begin{tabular}{c c c c c c} 
\hline
max\_age & min\_hits & min\_IoU & MOTA & IDEucl & IDF1 \\ [0.5ex]
\hline
1 & 3 & 0.3 & 41.1 & 28.4 & 30.3\\
\hline
\deemph{1} & \deemph{3} & 0.2 & 41.2 & 28.4 & 30.3\\

\deemph{1} & \deemph{3} & 0.4 & 41.0& 28.2 & 30.0\\
\hline

15 & \deemph{3} & \deemph{0.3} & 43.2 & 54.1 & 44.9\\

30 & \deemph{3} & \deemph{0.3} & 43.3 & 57.8 & 46.6\\

15 & 1 & \deemph{0.3} & 50.6 & 52.7 & 48.3\\

30 & 1 & \deemph{0.3} & 50.8 & 56.5 & 50.5\\

\deemph{1} & 1 & \deemph{0.3} & 46.8 & 27.3 & 30.5\\

\hline
\end{tabular}}
\end{center}
\caption{\label{tab:sort} Results depicting fine-tuning hyperparameter of SORT\cite{SORT} on the training set of CroHD.}
\end{table}

\subsection{HeadHunter-T}

We mainly analyze the impact of minimum confidence(or particle weights), $\lambda_{reg}$, required to keep a track alive. Table~\ref{tab:hht} shows the corresponding result. Surprisingly, lowering the $\lambda_{reg}$ performs the best amongst the other values. We believe thresholding detection to 0.6 to be a possible reason behind this observation. Hence, we also analyze the effect of $\lambda^{det}$, the minimum confidence score to initialize a track with $\lambda^{det}=0.8$ and $\lambda^{det}=0.3$. A reduction in $\lambda^{det}$ implied a mild deterioration in the identity preserving metrics, IDF1 and IDEucl. However, increasing $\lambda^{det}$ showed a noticeable decline in performance. An increment in the either initialization threshold ($\lambda^{det}$) or regression threshold ($\lambda_{reg}$) produces monotonically decreasing performance results. 
\begin{table}[htb]
\tabcolsep=0.05cm
\renewcommand{\arraystretch}{1.5}
\begin{center}
\resizebox{\linewidth}{!}{%
\begin{tabular}{c| c c c| c c c| c c c} 
\hline
\multirow{2}{*}{$\lambda_{reg}$} &
  \multicolumn{3}{c|}{$\lambda^{det}=0.3$} &
  \multicolumn{3}{c|}{$\lambda^{det}=0.6$} &
  \multicolumn{3}{c}{$\lambda^{det}=0.8$}\\
  \cline{2-10}
 & MOTA & IDEucl & IDF1 & MOTA & IDEucl & IDF1 & MOTA & IDEucl & IDF1\\ [0.5ex]
\hline
0.1 & \textbf{64.9} & 59.3 & 56.6 & 64.0 & \textbf{61.5} & \textbf{58.5} & 54.8 & 57.0 & 52.2\\

0.2 & 63.2 & 51.4 & 50.6 & 60.7 & 54.5 & 52.7 & 51.0 & 51.9 & 47.4\\

0.3 & 61.2 & 43.4 & 41.9 & 56.9 & 47.7 & 50.2 & 48.3 & 48.7 & 44.3\\

0.4 & 58.0 & 35.7 & 33.5 & 55.7 & 45.1 & 43.5 & 45.7 & 44.9 & 40.7\\

0.5 & 53.7 & 32.7 & 28.3 & 53.0 & 38.8 & 37.3 & 43.1 & 40.1 & 36.7\\

0.6 & 48.3 & 33.0 & 25.7 & 49.7 & 32.4 & 29.0 & 40.1 & 35.1 & 30.9\\


\hline
\end{tabular}}    
\end{center}
\caption{\label{tab:hht} Hyperparameter Fine-Tuning results of HeadHunter-T on the training set of CroHD.}
\end{table}

\subsection{Detection and Tracking}
In this section, we analyze the tracking performances of various object detectors that were used for baseline experiments on head detection task of CroHD. Table~\ref{tab:comp_det} shows the object detectors upon whose output, the initialization of tracks in HeadHunter-T depends on. The tracking performances were evaluated on the training set of CroHD. These experiments were preformed analogous to Public Detection experiments on the standard MOTChallenge Benchmarks~\cite{MOT19_CVPR,MOT16}. Since the task of Face Detection is cognate to Head Detection, we used RetinaFace~\cite{WIDER_FACE}, a recent face detector which is the state-of-the-art method on WIDER FACE dataset. We used the implementation and model weights provided by the author. HeadHunter without Fine-Tuning on CroHD and without the Context Module are denoted as HeadHunter W/O FT and HeadHunter W/O Ctx respectively. For Headhunter  W/O FT, we trained only on the training sets of CrowdHuman~\cite{shao2018crowdhuman} and SCUT-HEAD dataset~\cite{scut_head} . Barring RetinaFace and HeadHunter W/O FT, the remaining head detectors have been trained on CroHD. 

\begin{table}[htb]
\tabcolsep=0.01cm
\begin{center}
\renewcommand{\arraystretch}{1.5}
\resizebox{\columnwidth}{!}{
    \begin{tabular}{l c c c c c c}
        \hline
        Method & MOTA $\uparrow$ & IDEucl $\uparrow$ & IDF1 $\uparrow$ & MT $\uparrow$ & ML $\downarrow$ & ID Sw. $\downarrow$ \\ [0.5ex] 
        \hline
        FRCNN\cite{FRCNN} & 46.0 & 37.8 & 36.1 & 140 & 111 & 12,178\\
        
        FPN\cite{FPN}  & 49.1 & 37.0 & 35.5 & 202 & 95 & 10,424 \\ 
        
        HeadHunter W/O Ctx & 49.7 & 44.0 & 42.3 & 115 & 193 & 2,579 \\
        
        HeadHunter W/O FT & 54.5 & 40.0 & 38.4 & 142 & 116.0 &  7,621 \\
        
        RetinaFace\cite{retina_face} & 27.7 & 41.1 & 29.0 & 34.5 & 455 & 2,304\\
        \hline
        \textbf{HeadHunter-T} & \textbf{58.2} & \textbf{52.5} & \textbf{49.9} & \textbf{157} & \textbf{122} & \textbf{1941} \\
        \hline
    \end{tabular}}
\end{center} 
\caption{Tracking performance comparison of HeadHunter-T on training set of CroHD with tracked initialized from various detectors.}
\label{tab:comp_det}
\end{table}

\begin{table*}[]
\begin{center}
\renewcommand{\arraystretch}{1.5}
\resizebox{\linewidth}{!}{%
\begin{tabular}{|l|c|c|c|c|c|c|c|c|c|c|c|c|c|}
\hline
\multirow{2}{*}{Sequence Name} &
  Avg Track Length  &
  Avg Track Duration &
  Avg Velocity &
  \multicolumn{2}{c|}{BBRR} &
  Avg Occlusions &
  \multicolumn{4}{c|}{Instances per class} \\
  & (pixels) & (frames) & ($px.s^{-1}$)&width&height& (frames) &1&2&3&4 \\ \hline
CroHD-01   & 593 & 244.3 & 61.7  & 1:41.7 & 1:82.2 & 11.8 & 79 & 4 & 2 & 0    \\ 
CroHD-02   & 889 & 533.4 & 41.7 & 1:43.2 & 1:75.00 & 12.2 & 1,249 & 22 & 2 & 3  \\
CroHD-03 & 1,322 & 318.1 & 103.9 & 1:33.1 & 1:63.4 & 25.7 & 809 & 0 & 0 & 2    \\ 
CroHD-04 & 625 & 294.1  & 53.2 & 1:32.4 & 1:58.0 & 24.2 & 573 & 7 & 0 & 0    \\ \hline
CroHD-11   & 613 & 270.0  & 56.8 & 1:36.6 & 1:79.7 & 16.9 & 120 & 9 & 2 & 2    \\ 
CroHD-12   & 1,043 & 454.7 & 57.3 & 1:30.9 & 1:59.9 & 11.9  & 708 & 28 & 0 & 1   \\ 
CroHD-13 & 922 & 351.7 & 65.5 & 1:32.7 & 1:68.0 & 53.3 & 731 & 2 & 1 & 0    \\ 
CroHD-14 & 523 & 381.1 & 34.3 & 1:43.6 & 1:82.9 & 27.3  & 527 & 35 & 478 & 0 \\ 
CroHD-15 & 919 & 389.6  & 59.0 & 1:32.9 & 1:84.6 & 25.8 & 256 & 61 & 1 & 3\\ 
\hline
\end{tabular}%
}    
\end{center}
\caption{\label{tab:benchmark_robust_stat} Detailed statistics of each sequence composing our dataset, CroHD. BBRR indicates bounding box to image ratio (in pixels). Classes correspond to 1:Pedestrian, 2:Static, 3:Ignore and 4:Person on Vehicle.}
\end{table*}

\begin{table*}[htb]
\begin{center}

\renewcommand{\arraystretch}{1.5}
\resizebox{\linewidth}{!}{%
\begin{tabular}{c| c c c c c c|c c c c c c c c} 
\hline
\multirow{2}{*}{Sequence Name} &
  \multicolumn{6}{c|}{Head Detection} &
  \multicolumn{8}{c}{Head Tracking}\\
\cline{2-14}
& AP $\uparrow$ & R $\uparrow$ & F1 $\uparrow$ & MODA $\uparrow$ & MODP $\uparrow$ & mAP\_COCO $\uparrow$ & MOTA  $\uparrow$ & IDF1 $\uparrow$ & IDEucl $\uparrow$ & MT $\uparrow$ & ML $\downarrow$ & FP $\downarrow$ & FN $\downarrow$ & IDs $\downarrow$\\ [0.5ex]
\hline
CroHD-01 & 79.3 & 83.4 & 86.5 & 76.4 & 64.0 & 37.3 & 84.5 & 76.4 & 79.1 & 55 & 4 & 237 & 2,550 & 59 \\

CroHD-02 & 40.4 & 52.9 & 61.1 & 50.0 & 38.6 & 9.1 & 66.7 & 66.4 & 60.0 & 548 & 127 & 46,479 & 168,299 & 2,049 \\

CroHD-03 & 58.9 & 60.4 & 73.3 & 61.6 & 45.5 & 17.2 & 51.3 & 45.4 & 42.9 & 160 & 133 & 9,481 & 103,562 & 2,243  \\

CroHD-04 & 64.6 & 70.0 & 76.9 & 65.7 & 51.5 & 20.3 & 53.6 & 52.7 & 47.9 & 135 & 98 & 9,438 & 61,238 & 975 \\
\hline
CroHD-11 & 83.1 & 86.4 & 88.3 & 79.5 & 64.9 & 37.4 & 81.5 & 76.1 & 75.2 & 84 & 7 & 1,428 & 4,056 & 101 \\

CroHD-12 & 34.8 & 51.0 & 58.6 & 42.1 & 37.2 & 10.2 & 60.6 & 64.3 & 57.1 & 264 & 64 & 21,851 & 100,484 & 1,173 \\

CroHD-13 & 41.7 & 45.6 & 58.8 & 47.0 & 32.6 & 11.1 & 32.5 & 29.5 & 28.1 & 29 & 296 & 11,499 & 133,789 & 2,034 \\

CroHD-14 & 45.8 & 62.3 & 67.5 & 43.1 & 46.7 & 16.0 & 67.3 & 61.2 & 59.4 & 215 & 60 & 11,506 & 48,580 & 817\\

CroHD-15 & 57.5 & 71.8 & 68.5 & 38.7 & 54.9 & 24.2 & 75.9 & 70.4 & 65.9 & 140 & 76 & 5,540 & 16,710 & 334 \\
\hline
\end{tabular}}
\end{center}
\caption{\label{tab:tracking_ourbm} Sequence-wise performances of HeadHunter and HeadHunter-T on CroHD.}
\end{table*}

\begin{algorithm}[h]
\caption{HeadHunter-T}\label{algorithm2}
	\begin{algorithmic}[1]
	     \REQUIRE $\mathrm{Video}\ \mathcal{I}\ \ \mathrm{containing\ T\ frames}\ \{\mathcal{I_{0}},\cdots,\mathcal{I_{T-1}}\}$
	     \ENSURE $\mathrm{Trajectories}\ \mathcal {T}=\left\{\mathcal{T}_{1}, \cdots, \mathcal{T}_{k}\right\}$
	    \STATE $\mathcal{L}, \mathcal{T}, \mathcal{D} \leftarrow \phi$
        \FOR {$t=1, \cdots, T-1$}
            \STATE $\textbf{F}_{t} \leftarrow \featext(\mathcal{I}_{t})$
            \FOR {$l \in \mathcal{L} $}
                \IF {$l.\lambda^{t} > \lambda^{age}$}
                    \STATE $\mathcal{L}_{t} \leftarrow \mathcal{L}_{t} \setminus l$
                \ENDIF
                \STATE $l.\mathrm{predict\_cva()}$
            \ENDFOR
            \FOR {$a \in \mathcal{T} $}
                \STATE $\overline{\textbf{p}_{t}^{a}}, \overline{\textbf{w}_{t}^{a}} \leftarrow \roipool(\textbf{F}_{t}, \overline{\mathbf{p}}_{t-1}^{a} \mathrm{.predict())}$
                \IF{$\mathrm{mean}(\overline{\mathrm{\textbf{w}}_{t}^{a}}) < \lambda^{\mathrm{reg}}$}
                    \STATE $\mathcal{T} \leftarrow \mathcal{T} \setminus a$
                    \STATE $\mathcal{L} \leftarrow \mathcal{L} \cup a$
                \ELSE
                    \STATE $\mathcal{T}\cup a$
                \ENDIF
                \IF{$\hat{\mathbf{N}}_{\mathrm{eff}}^{k} > \hat{\mathbf{N}}_{\mathrm{thresh}}$}
                    \STATE $\mathrm{a.resample}(\hat{\mathbf{p}}^{a}_{t})$
                \ENDIF
            \ENDFOR
            \STATE $\mathcal{D}_{t} \leftarrow \mathrm{filter}(\roipool(\rpn(F_{t})), \lambda^{\mathrm{new}}) $
            \STATE $ \mathcal{D}_{t} \leftarrow \mathcal{D}_{t} \setminus \mathrm{filter}(\mathrm{IoU}(\mathcal{D}_{t}, \mathcal{T}_{t}), \lambda^{\mathrm{init}})$
            \FOR{$d \in \mathcal{D}_{t}$}
                \FOR {$l \in \mathcal{L} $}
                    \IF{$\mathrm{cost\_match}(l, d, \alpha, \beta) > \mathcal{C}$}
                        \STATE $\mathcal{L}_{t} \leftarrow \mathcal{L}_{t} \setminus l$
                        \STATE $\mathcal{D}_{t} \leftarrow \mathcal{D}_{t} \setminus l$
                        \STATE $\mathcal{T} \leftarrow \mathcal{T} \cup l$
                        \STATE $\mathrm{init\_particles}(l)$
                    \ENDIF
                \ENDFOR
            \ENDFOR
            \FOR{$d \in \mathcal{D}_{t}$}
                \STATE $\mathcal{N} \leftarrow \mathrm{init\_particles}(\mathrm{init\_track}(d))$
            \ENDFOR
        \STATE $\mathcal{T} \leftarrow \mathcal{T} \cup \mathcal{N} \And \mathcal{N} \leftarrow \phi$
        \ENDFOR
        \RETURN{} $\mathcal{T}$
	\end{algorithmic} 
\end{algorithm}

{\clearpage
\small
\bibliographystyle{ieee_fullname}
\bibliography{egbib}
}